%% file: main.tex
\documentclass[journal]{IEEEtran}
\usepackage[noadjust]{cite}
\usepackage{times}
\usepackage{amsthm}
\usepackage{amsmath}
\usepackage{amsfonts}
\usepackage{amssymb}
\usepackage{float}
\usepackage{stfloats}
\usepackage{graphicx}
\usepackage[caption=false,font=footnotesize,labelfont=sf,textfont=sf]{subfig}
\usepackage{multicol}
\usepackage{multirow}
\usepackage[bookmarks=true]{hyperref}
\usepackage{nicefrac}       
\usepackage{microtype}      
\usepackage{algorithm}
\usepackage[noend]{algpseudocode}
\usepackage[table]{xcolor}
\usepackage{flushend}

\algnewcommand{\LineComment}[1]{\State \(\triangleright\) #1}

\input{terms.tex}

\title{Constrained Stein Variational Trajectory Optimization}

\author{Thomas Power$^{1}$ and Dmitry Berenson$^{1}$
\thanks{This work was sponsored by Honda Research Institute USA. $^1$Authors are with the Robotics Department, University of Michigan, Ann Arbor, MI, USA. {\tt\small \{tpower, dmitryb\}@umich.edu}}%
}
\begin{document}
\maketitle

\begin{abstract}
We present Constrained Stein Variational Trajectory Optimization (CSVTO), an algorithm for performing trajectory optimization with constraints on a set of trajectories in parallel. We frame constrained trajectory optimization as a novel form of constrained functional minimization over trajectory distributions, which avoids treating the constraints as a penalty in the objective and allows us to generate diverse sets of constraint-satisfying trajectories. Our method uses Stein Variational Gradient Descent (SVGD) to find a set of particles that approximates a distribution over low-cost trajectories while obeying constraints. CSVTO is applicable to problems with differentiable equality and inequality constraints and includes a novel particle re-sampling step to escape local minima. By explicitly generating diverse sets of trajectories, CSVTO is better able to avoid poor local minima and is more robust to initialization. We demonstrate that CSVTO outperforms baselines in challenging highly-constrained tasks, such as a 7DoF wrench manipulation task, where CSVTO outperforms all baselines both in success and constraint satisfaction.


\end{abstract}

\section{Introduction}
\IEEEPARstart{T}{rajectory} optimization and optimal control are powerful tools for synthesizing complex robot behavior using appropriate cost functions and constraints \cite{gusto, trajopt, altro, ipopt_trajectory2, mpc_bipeds}.  Constraint satisfaction is important for safety-critical applications, such as autonomous driving, where constraints determine which trajectories are safe. Constraints can also provide effective descriptions of desired behavior. For instance, consider a robot sanding a table. This problem can be defined with an equality constraint specifying that the end-effector must move along the surface of the table as well as constraints on the minimum and maximum force applied to the table. For many tasks, including manipulation tasks like the one above, satisfying these constraints can be very difficult as constraint-satisfying trajectories may lie on implicitly defined lower-dimensional manifolds. Such constraints present difficulties for sample-based methods since the feasible set has zero measure and thus it is difficult to sample. It is also difficult for gradient-based methods since even for trajectories that start feasible, if the constraint is highly nonlinear then updates based on a first-order approximation of the constraint will lead to solutions leaving the constraint manifold. In addition, many useful tasks entail constrained optimization problems that are non-convex and exhibit multiple local minima. 

Global sample-based motion planning methods such as Rapidly-Exploring-Random-Trees (RRT) \cite{rrt_connect}. Probabilistic Roadmaps (PRM) \cite{prm} effectively solve difficult planning problems, however, they do not find paths that minimize a given cost function. To minimize a given cost function, algorithms such as RRT* and PRM* \cite{rrtstar} have been proposed to find asymptotically globally optimal paths. Alternatively, a common approach is to use the path returned from a sample-based motion planner to initialize a trajectory optimization problem \cite{lavalle:2006}. Sample-based methods have additionally been applied to constrained planning problems \cite{tsr, projection_mp_2, constrained_planning, continuation}, and kinodynamic problems \cite{sst, kinodynamic_rrt}. While effective for solving problems exhibiting local minima, when applied to kinodynamic or constrained problems these global methods are typically computationally expensive. 

\begin{figure}
    \centering
    \includegraphics[width=0.48\textwidth]{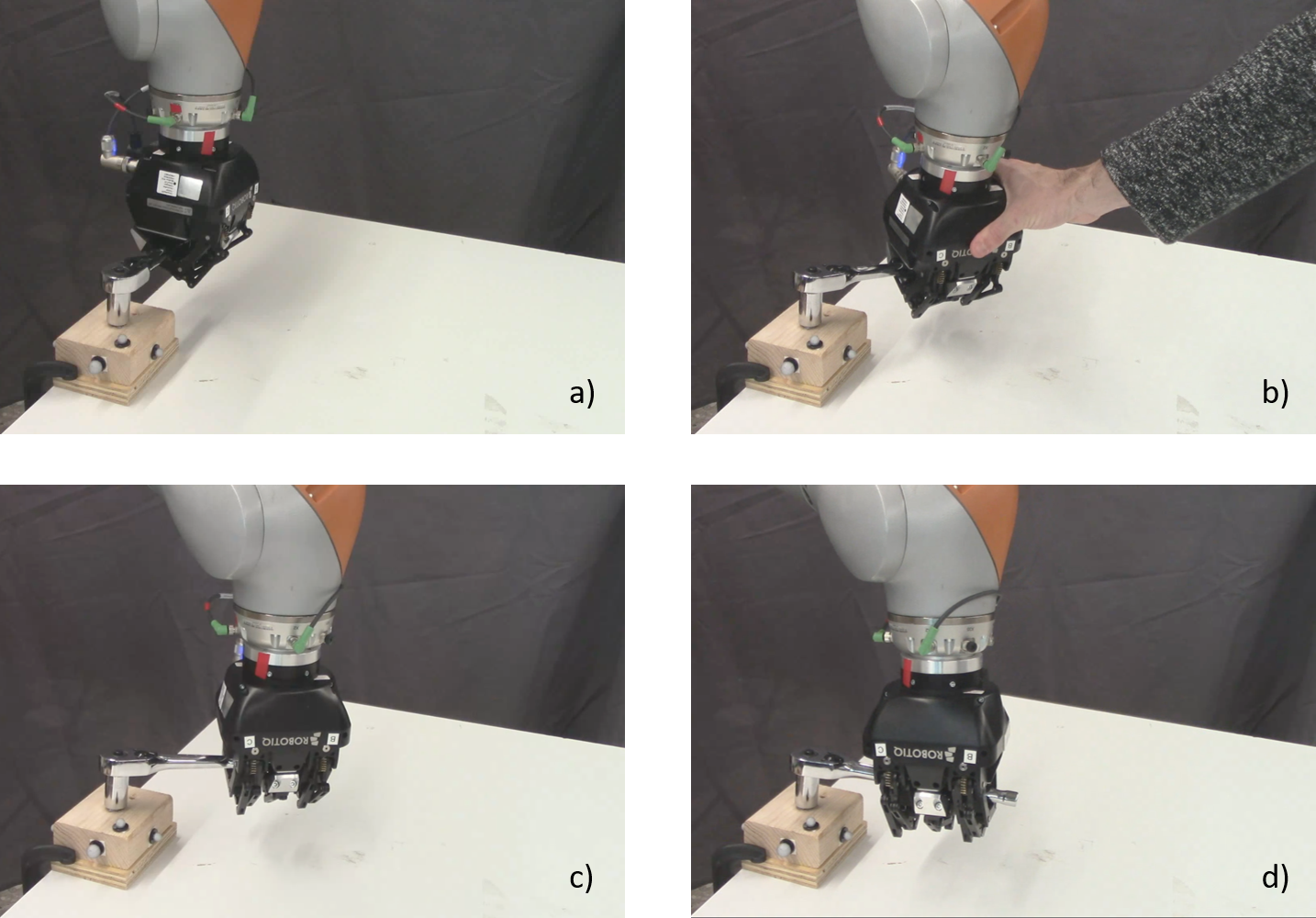}
    \caption{We use CSVTO to turn a wrench in the real world with online replanning; b) A human disturbs the robot, changing the grasp position of the wrench; c) The robot readjusts the grasp position; d) The robot achieves the desired wrench angle.} 
    \label{fig:victor_real}
\end{figure}
One of the key advantages of trajectory optimization techniques over global search methods, such as sampling-based motion planning, is computation speed. Faster computation speed enables online re-planning to adapt to disturbances. For example, consider again the robot sanding the table, but now in the proximity of a human. The human may move in an unexpected way which necessitates an update to the planned trajectory. However, even if the cost function is well-suited to the task, the performance of many trajectory optimization methods is still highly dependent on the initialization. Poor initialization may lead to the solver converging to a poor local minimum. For example, for a robot minimizing a distance to goal cost subject to collision constraints, this may mean a trajectory that avoids obstacles but makes little or no progress toward the goal. In the worst case, the solver may not find a feasible solution, in which case the robot may collide with an obstacle. A dependency on initialization is particularly problematic when re-solving the optimization problem online under limited computation time when disturbances can lead to the previous solution becoming a poor initialization for the current optimization problem. In the sanding example mentioned previously, the human may move to block the robot's path, and performing a local optimization starting from the previous trajectory may not return a feasible solution. 

In this article, we formulate the constrained trajectory optimization problem as a Bayesian inference problem. This view has advantages as it aims to find a distribution over trajectories rather than a single trajectory alone. As noted by Lambert et. al. \cite{lambert2021entropy}, commonly used Variational Inference approaches \cite{VI} lead to minimizing entropy-regularized objectives \cite{lambert2021entropy} which can improve exploration of the search space and give greater robustness to initialization. Previous methods taking the inference view of trajectory optimization have only been able to incorporate constraints via penalties in the cost \cite{gp_variational_motion_planning, GPmotion_planning, lambert2021entropy, stein_mpc}. A drawback of penalty methods is that selecting the relative weights of the penalties is challenging due to possible conflicts with the objective. We compare against baselines that incorporate constraints via penalties and show that, for non-trivial constraints, this results in poor constraint satisfaction. An alternative method for enforcing constraints in trajectory optimization is via barrier functions \cite{barrier1, barrier2}. While effective, they are only applicable to inequality constraints and have not yet been applied in the context of trajectory optimization as an inference problem.

We propose Constrained Stein Variational Trajectory Optimization (CSVTO), an algorithm that performs constrained trajectory optimization on a set of trajectories in parallel. Our method builds on Orthogonal-Space Stein Variational Gradient Descent (O-SVGD), a recent non-parametric variational inference method for domains with a single equality constraint \cite{constrained_stein}. We present a constrained Stein Variational Gradient Descent (SVGD) algorithm for trajectory optimization with differentiable equality and inequality constraints, generating a diverse set of approximately constraint-satisfying trajectories. The trajectories are \textit{approximately} constraint-satisfying because we do not run the algorithm until convergence to avoid excessive computation times. We additionally incorporate a novel re-sampling step that re-samples and perturbs particles in the tangent space of the constraints to escape local minima. Our contributions are as follows:
 \begin{itemize}
     \item We frame constrained trajectory optimization as a novel form of constrained functional minimization over trajectory distributions, which avoids treating the constraints as a penalty in the objective. 
     \item We present a constrained SVGD algorithm for trajectory optimization, which is applicable to problems with differentiable equality and inequality constraints.
     \item We propose a novel particle re-sampling step for re-sampling and perturbing trajectory particles in the tangent space of the constraints to escape local minima. 
     \item We evaluate our method on three complex constrained problems, including a 12DoF underactuated quadrotor and two highly constrained 7DoF manipulation tasks. 
 \end{itemize}

Our experimental results demonstrate that CSVTO outperforms baselines in challenging, highly constrained tasks, such as a 7DoF wrench manipulation task where our method achieves 20/20 success compared with 12/20 for Interior Point OPTimizer (IPOPT) \cite{ipopt} and 19/20 for Stein Variational Model Predictive Control (SVMPC) \cite{stein_mpc}, CSVTO also achieves the lowest constraint violation of all baselines. In addition, CSVTO outperforms baselines in a 12DoF quadrotor task with a dynamic obstacle that necessitates online adaption of the planned trajectory.

The rest of the article is organized as follows. In Section \ref{sec:related_work} we discuss related work. In Section \ref{sec:trajopt} we will discuss the trajectory optimization problem, followed by an overview of the variational inference approach to trajectory optimization in Section \ref{sec:vi_trajopt}. In Section \ref{sec:problem_statement} we introduce our novel formulation of trajectory optimization as a constrained functional minimization over trajectory distributions. We will then give some additional background information on SVGD in  Section \ref{sec:stein} which is necessary to develop our algorithm. In Section \ref{sec:methods} we introduce CSVTO. In Section \ref{sec:eval} we evaluate our method on a 12DoF quadrotor task and two highly constrained tasks with a 7DoF manipulator. We additionally deployed CSVTO to turn a wrench in the real world (Figure \ref{fig:victor_real}). 

\section{Related Work}
\label{sec:related_work}
\subsection{Trajectory optimization}
Previous work on local trajectory optimization techniques includes direct methods \cite{ipopt_trajectory1, ipopt_trajectory2}, where the explicit optimization problem is transcribed and solved using nonlinear solvers such as IPOPT \cite{ipopt} or Sparse Nonlinear OPTimizer (SNOPT) \cite{snopt}. Methods in this class include Sequential Convex Programming (SCP) methods such as TrajOpt \cite{trajopt} and Guaranteed Sequential Trajectory Optimization (GuSTO) \cite{gusto}. In contrast, indirect methods aim instead to solve the local optimality conditions of the trajectory and early examples include Differential Dynamic Programming (DDP) \cite{DDP} and iterative Linear Quadratic Regulator (iLQR) \cite{iLQR}, however, neither of these methods can handle constraints. Later work incorporated constraint satisfaction with these indirect methods \cite{ilqr_constr, ddp_constr, altro}. Direct methods are typically easier to initialize but less accurate {\cite{von1992direct}. However all of these methods only aim to find a single locally-optimal trajectory, and the performance is dependent on the initialization. In contrast, our approach optimizes a diverse set of trajectories in parallel. This makes our approach easier to initialize as well as more robust to disturbances when re-planning online. Our approach is related to the direct methods, in that we use an iterative algorithm that aims to minimize an objective. However, our method is based on viewing the trajectory optimization problem as a Bayesian inference problem.

\subsection{Sample-based Motion Planning}
Many global search methods have been developed in the sampling-based motion planning literature, yielding motion planners for constrained domains. These can be broadly categorized as projection methods, whereby sampled configurations are projected to the constraint \cite{tsr, projection_mp_2}, and continuation methods, which use a local approximation of the constraint manifold at feasible configurations to sample new configurations \cite{constrained_planning, continuation}. Our method of trajectory optimization is similar to continuation methods, as our iterative algorithm projects update steps to the tangent space of the constraint. While these global motion planners can be highly effective, they are typically too computationally intensive to be run online. 

\subsection{Planning \& Control as Inference}
Prior work framing trajectory optimization as Bayesian inference has used Gaussian approximations to yield fast, gradient-based algorithms \cite{GPmotion_planning, Gp_mp2, toussaint_probabilistic_2006, soc_as_inference, watson_socinf, gp_variational_motion_planning}. Ha et al. presented a probabilistic approach for trajectory optimization with constraints, using Laplace approximations around local minima found by solving a non-linear program (NLP) \cite{ha2020probabilistic}. This approach uses a Gaussian approximation with a degenerate covariance with variance only in the tangent space of the constraints. Samples from this distribution will generally deviate from the constraint manifold for non-linear constraints, in contrast, our approach directly optimizes for diverse constraint-satisfying samples.} Sample-based techniques such as Model Predictive Path Intregral (MPPI) control \cite{mppi} and Cross-Entropy Method (CEM) \cite{CEM} have strong connections to the inference formulation of Stochastic Optimal Control (SOC) \cite{VI_Tsallis}, but these methods again use Gaussian sampling distributions.  Several recent works have focused on improving the performance of these algorithms, often by modifying the sampling distribution. Watson and Peters recently proposed using a Gaussian Process as a sampling distribution \cite{gp_mpc}, and Pinneri et al. proposed using colored noise \cite{iCEM}, both of which lead to smoother sampled trajectories. Bhardwaj et al. \cite{storm} has also demonstrated improvements to MPPI with a focus on robot manipulation. However, in all of these prior works, the sampling distribution is uni-modal. Uni-modal sampling distributions can be problematic in complex environments due to their lack of flexibility which hinders exploration of the search space. Recent work has proposed learning non-Gaussian sampling distributions with flexible model classes \cite{FlowMPPI, sacks2023learning}. 

Another class of methods has used Stein Variational Gradient Descent (SVGD) \cite{liu2016stein} for Model Predictive Control (MPC) \cite{stein_mpc, dual_stein} and trajectory optimization \cite{lambert2021entropy}. By using particle approximations these methods can generate multi-modal trajectory distributions. SVGD has also been used to improve Probabilistic Roadmaps (PRMs) \cite{stein_roadmaps}, and for planning to goal sets \cite{pmlr-v229-pavlasek23a}. Our method is also based on SVGD. 

However, to date, control-as-inference-based methods have been unable to handle highly constrained domains. Recently Constrained Covariance Steering MPPI (CCSMPPI) \cite{constrained_mppi} was proposed which can satisfy chance inequality constraints, but is restricted to linear systems. Our method uses SVGD to generate diverse sets of \textit{constraint-satisfying} trajectories which can satisfy both inequality and equality constraints. Another method closely related to ours is Stochastic Multimodal
Trajectory Optimization (SMTO) \cite{multimodal_trajopt}, this method treats the trajectory optimization problem as a density estimation problem and alternates between sampling and performing a gradient-based optimization to generate multiple low-cost trajectories that satisfy the constraints. SMTO uses Covariant Hamiltonian Optimization
for Motion Planning (CHOMP) \cite{chomp} to perform the gradient-based optimization sequentially for each sampled trajectory. Our contribution is complementary to SMTO; SMTO could substitute CHOMP with our method, CSVTO, in the gradient-optimization step. This would have the advantage of performing the gradient-based optimization in parallel and encouraging diversity among trajectories. 

\subsection{Gradient Flows for constrained optimization} 
Our method is closely related to methods using gradient flows for constrained optimization. Gradient flows are an optimization method that re-frames optimization as the solution to an ordinary differential equation (ODE); gradient flows can be thought of as continuous-time versions of gradient descent algorithms. Yamashita proposed a gradient flow method for equality-constrained problems \cite{yamashita1980differential}. The most common method of extending this to problems with inequality constrained is via the introduction of slack variables to convert inequality constraints to equality constraints \cite{ineq_flow, projected_gradient_flow, constrained_global_flow}. Our method, CSVTO, also uses slack variables to transform inequality constraints into equality constraints. Recently, Feppon et. al. \cite{feppon2020optim} proposed a method that instead solves a Quadratic Program (QP) subproblem to identify active inequality constraints which are treated as equality constraints in the gradient flow. Jongen and Stein applied constrained gradient flows to global optimization, by proposing a gradient flow algorithm that iterates between searching for local minima and local maxima \cite{constrained_global_flow}. 

SVGD has been interpreted as a gradient flow \cite{svgd_flow}, and similar ideas to those developed in the gradient flows for constrained optimization literature were recently explored in O-SVGD \cite{constrained_stein}. O-SVGD performs SVGD in domains with a single equality constraint. We extend and modify O-SVGD to domains with multiple equality and inequality constraints.

\section{Trajectory Optimization}
\label{sec:trajopt}
Trajectory optimization is commonly modeled as an Optimal Control Problem (OCP). We consider a discrete-time system with state $\mathbf{x} \in \mathbb{R}^{d_x}$ and control $\mathbf{u} \in \mathbb{R}^{d_u}$, where $d_x$ and $d_u$ are the dimensionality of the state and control, respectively, and dynamics $\mathbf{x}_{t} = f(\mathbf{x}_{t-1}, \mathbf{u}_{t-1})$. We define finite horizon trajectories with horizon $T$ as $\tau = (\mathbf{X}, \mathbf{U})$, where $\mathbf{X} = \{\mathbf{x}_1, ... \mathbf{x}_T\}$ and $\mathbf{U} = \{\mathbf{u}_0, ... \mathbf{u}_{T-1}\}$. 
Given an initial state $\mathbf{x}_0$, the aim when solving an OCP is to find a trajectory $\tau$ that minimizes a given cost function $C$ subject to equality and inequality constraints: 
\begin{equation}
\begin{aligned}
\min_\tau & \quad C(\tau)\\
\textrm{s.t.} \\
  & h(\tau) = 0\\
  & g(\tau) \leq 0   \\
  & \forall t \in \{1, \dots, T\} \\
  & f(\mathbf{x}_{t-1}, \mathbf{u}_{t-1}) = \mathbf{x}_{t} \\
  & \mathbf{u}_\text{min} \leq \mathbf{u}_{t-1} \leq \mathbf{u}_{\text{max}} \\
  & \mathbf{x}_\text{min} \leq \mathbf{x}_t \leq \mathbf{x}_{\text{max}}. \\
\end{aligned}
\label{eq:trajectory_opt}
\end{equation}
Here we have separated general inequality constraints $g$ from simple bounds constraints, as well as the dynamics constraints from other equality constraints $h$. We additionally assume that $C$ is non-negative and once differentiable and that $f, g, h$ are all twice differentiable\footnote{We can also accommodate constraints that are only once-differentiable via an approximation (see Section \ref{sec:method:repulsion}}. Problem (\ref{eq:trajectory_opt}) will be non-convex in general, therefore it is likely it will have multiple local minima. The quality of solutions for most methods for solving this optimal control problem depends heavily on the initialization; often a poor initialization can lead to infeasibility.

\section{Variational Inference for Trajectory Optimization}
\label{sec:vi_trajopt}
In this section, we will demonstrate how unconstrained trajectory optimization can be framed as an inference problem, as in \cite{soc_as_inference, soc_as_inference_2, okada_variational_mpc, stein_mpc}. This framing results in estimating a distribution over low-cost trajectories, rather than a single optimal trajectory. By using this framing we can leverage approximate inference tools for trajectory optimization, in particular, Variational Inference \cite{VI}. In this section, we will show how this framing leads to an entropy-regularized objective \cite{lambert2021entropy} which aims to find a distribution over low-cost trajectories while maximizing entropy. By using an entropy-regularized objective we aim to have improved exploration of the search space and greater robustness to initialization. 

To reframe trajectory optimization as probabilistic inference, we first introduce an auxiliary binary random variable $o$ for a trajectory such that
\begin{equation}
    p(o=1|\tau) = \exp{(- \gamma C(\tau))},
\end{equation}
which defines a valid probability distribution over $o$ provided both $\gamma$ and $C$ are non-negative. We can trivially see that the trajectory that maximizes the likelihood of $p(o=1|\tau)$ is the trajectory that minimizes the cost. Introducing this binary variable allows us to express the cost as a likelihood function, which we will use in the Bayesian inference formulation of trajectory optimization. Using this likelihood to perform inference gives us a principled way of computing a distribution over trajectories, where lower-cost trajectories have a higher likelihood. The term $\gamma$ controls how peaked the likelihood function is around local maxima, or minima of $C$, which in turn controls the dispersion of the resulting trajectory distribution after performing inference.

We aim to find the posterior distribution over trajectories, conditioned on the value of auxiliary variable $o$. This is given by Bayes theorem as
\begin{equation}
    p(\tau | o=1) = \frac{p(o=1 | \tau) p(\tau)}{p(o=1)},
\end{equation}

\noindent where $p(\tau) = p(\mathbf X, \mathbf U)$ is a prior on trajectories. For deterministic dynamics, this prior is determined by placing a prior on controls $\mathbf U$. This prior is a design choice and can be used to regularize the controls. For instance, a squared control cost can be equivalently expressed as a Gaussian prior. Alternatively, this prior could be learned from a dataset of low-cost trajectories \cite{learning_priors}. The trajectory prior is
\begin{equation}
p(\tau) = p( \mathbf U) \prod ^T_{t=1} \delta(\mathbf{x}_t -\mathbf{\hat{x}}_t),
\label{eq:prior}
\end{equation}
where $\mathbf{\hat{x}}_t = f(\mathbf{x}_{t-1}, \mathbf{u}_{t-1})$, and $\delta$ is the Dirac delta function. This inference problem can be performed exactly for the case of linear dynamics and quadratic costs \cite{planning_as_inference, watson_socinf}. However, in general, this problem is intractable and approximate inference techniques must be used. We use variational inference to approximate  $p(\tau | o=1)$ with distribution $q(\tau)$ which minimizes the Kullback–Leibler divergence  $\mathcal{KL}(q(\tau) || p(\tau | o = 1))$ \cite{VI}. The quantity to be minimized is
\begin{align}
\begin{split}
&\mathcal{KL}\left(q (\tau) || p(\tau | o=1) \right) = \int q(\tau) \log \frac{q(\tau)}{p(\tau | o=1)} d\tau \\
&= \int q(\tau) \log \frac{q(\tau)p(o=1)}{p(o=1|\tau)p(\tau)}d\tau. 
\end{split}
\end{align}
The $p(o=1)$ term in the numerator does not depend on $\tau$ so can be dropped from the minimization. This results in the \textit{variational free energy} $\mathcal{F}$:
\begin{align}
\mathcal{F}(q) &= \int q(\tau) \log \frac{q(\tau)}{p(o=1|\tau)p(\tau)}d\tau \\
\begin{split}
&= -\mathbb{E}_{q(\tau)}[\log p(o=1| \tau) + \log p(\tau)] - \mathcal{H}(q(\tau))
\label{background:free_energy} \\
\end{split} \\
&= \mathbb{E}_{q(\tau)}[\gamma C(\tau)] - \mathbb{E}_{q(\tau)} [\log p(\tau)] - \mathcal{H}(q(\tau)), \label{app:objective}
\end{align}
where $\mathcal{H}(q(\tau))$ is the entropy of $q(\tau)$. Intuitively, we can understand that the first term promotes low-cost trajectories, the second is a regularization on the trajectory, and the entropy term prevents the variational posterior from collapsing to a \textit{maximum a posteriori} (MAP) solution. We may choose to provide regularization on the controls as part of $C$, in which case the prior term is absorbed into the cost term. 

\section{Problem Statement}
\label{sec:problem_statement}
In this article, we frame the constrained optimal control problem introduced in Section \ref{sec:trajopt} as a probabilistic inference problem, using ideas developed in Section \ref{sec:vi_trajopt}. 

It is first instructive to consider the dynamics constraint, which is incorporated into the prior in equation (\ref{eq:prior}) via the Dirac delta function. In this case, the term $\mathbb{E}_{q(\tau)}[-\log p(\tau)]$ is infinite for any $\tau$ which does not obey the dynamics constraint. We can convert this unconstrained optimization problem with infinite cost to the following constrained optimization problem on the space of probability distributions:
\begin{equation}
\begin{aligned}
\min_{q} & \quad \mathcal{\tilde{F}}(q) \\
\textrm{s.t.} \\
  & \forall t \in \{1, \dots, T\} \\
  & P_q(f(\mathbf{x}_{t-1}, \mathbf{u}_{t-1}) = \mathbf{x}_{t}) = 1,
\end{aligned}
\end{equation}
where $\mathcal{\tilde{F}}(q)$ is the free energy from equation (\ref{app:objective}) with the infinite cost term $\sum^T_{t=1} \log \delta(\mathbf{x}_{t} - f(\mathbf{x}_{t-1}, \mathbf{u}_{t-1}))$ dropped from $\log p(\tau)$, and $P_q(A)$ is the probability of event $A$ under probability measure $q(\tau)$. Applying this process to other constraints we have
\begin{equation}
\begin{aligned}
\min_{q} & \quad \mathcal{\tilde{F}}(q) \\
\textrm{s.t.} \\
  & P_q(h(\tau) = 0) = 1\\
  & P_q(g(\tau) \leq 0) = 1   \\
  & \forall t \in \{1, \dots, T\} \\
  & P_q(f(\mathbf{x}_{t-1}, \mathbf{u}_{t-1}) = \mathbf{x}_{t}) = 1 \\
  & P_q(\mathbf{u}_\text{min} \leq \mathbf{u}_{t-1} \leq \mathbf{u}_{\text{max}}) = 1 \\
  & P_q(\mathbf{x}_\text{min} \leq \mathbf{x}_t \leq \mathbf{x}_{\text{max}}) = 1.\\
  \label{eq:constrained_planning_inference}
\end{aligned}
\end{equation}
Our goal is to solve the above optimization problem. However, for any practical algorithm, we cannot guarantee exact constraint satisfaction, both due to the potential non-convexity of the constraint functions and due to limited computation time. Computation time is especially limited in an online planning scenario. Therefore we will evaluate our method according to both the cost of the resulting trajectories and the amount of constraint violation when optimizing within a fixed number of iterations. 
\section{Stein Variational Gradient Descent}
\label{sec:stein}
We develop an algorithm to solve the constrained variational inference objective in (\ref{eq:constrained_planning_inference}) based on Stein Variational Gradient Descent (SVGD) \cite{liu2016stein}. In this section we will give an overview of SVGD which forms the foundation of our method. 
SVGD is a variational inference technique that uses a non-parametric representation of the variational posterior. In our algorithm, we use SVGD to approximate the distribution $p(\tau|o=1)$ with particles, where each particle is a trajectory. Consider the variational inference problem
\begin{equation}
    q^*(\mathbf{x}) = \arg \min_{q(\mathbf{x})} \quad \mathcal{KL}\left(q(\mathbf{x}) || p(\mathbf{x}) \right),
\end{equation}
where $\mathbf x \in \mathbb{R}^d$ and $p$ and $q$ are two probability density functions supported on $\mathbb{R}^d$. SVGD uses a particle representation of $q(\mathbf{x}) = \frac{1}{N} \sum^N_{i=1}\delta (\mathbf{x} - \mathbf{x}^i)$, and iteratively updates these particles in order to minimize $\mathcal{KL}\left(q(\mathbf{x}) || p(\mathbf{x}) \right)$. SVGD updates the particle set with the update equation
\begin{equation}
    \mathbf{x}^i_{k+1} = \mathbf{x}^i_k + \epsilon \mathbf{\phi}^*(\mathbf{x}^i_k),
    \label{eq:stein1}
\end{equation}
where $\epsilon > 0$ is a step-size parameter, $k$ is the iteration number, and $i$ is the particle index. The update $\phi^*$ is computed using a differentiable positive definite kernel function $\mathcal{K}$ via
\begin{equation}
    \mathbf \phi^*(\mathbf{x}_k^i) = \frac{1}{N}\sum^N_{j=1} \mathcal{K}(\mathbf{x}^i_k, \mathbf{x}_k^j) \nabla_{\mathbf{x}_k^j} \log p(\mathbf{x}^j_k) + \nabla_{\mathbf{x}_k^j} \mathcal{K}(\mathbf{x}_k^i, \mathbf{x}_k^j).
    \label{eq:stein_update}
\end{equation}
The first term of this objective maximizes the log probability $p(\mathbf{x})$ for the particles, with particles sharing gradients according to their similarity defined by $\mathcal{K}$. The second term is a repulsive term that acts to push particles away from one another and prevents the particle set from collapsing to a local MAP solution.

We will now give further details on the derivation of the SVGD algorithm and demonstrate that it does indeed minimize $\mathcal{KL}\left(q(\mathbf{x}) || p(\mathbf{x}) \right)$. We will use the developments in this section to show that the fixed points of our algorithm satisfy first-order optimality conditions in section \ref{methods:analysis}.
SVGD is based on the \textit{Kernelized Stein Discrepancy} (KSD) \cite{ksd}, which is a measure of the discrepancy between two distributions $p$ and $q$. The KSD is computed as the result of the following constrained functional maximization
\begin{equation}
        \mathbb{S}(p, q) = \max_{\phi \in \mathcal{H}^d}  \left \{ \mathbb{E}_{\mathbf x \sim q}[\mathcal{A}_p \phi(\mathbf x)] \: \text{s.t.} \: ||\phi||_{\mathcal{H}^d} \leq 1 \right \},
        \label{eq:ksd}
\end{equation}
where $\phi : \mathbb{R}^d \to \mathbb{R}^d$ is a function in a vector-valued Reproducing Kernel Hilbert Space (RKHS) $\mathcal{H}^d$ with a scalar kernel $\mathcal{K} : \mathbb{R}^d \times \mathbb{R}^d \to \mathbb{R}$. $\mathcal{A}_p$ is the Stein operator
\begin{equation}
    \mathcal{A}_p \phi(\mathbf x) = \nabla_\mathbf{x} \log p(\mathbf x)^T \phi (\mathbf x) + \nabla_\mathbf{x} \cdot \phi(\mathbf x),
    \label{eq:stein_op}
\end{equation}
where $\nabla_{\mathbf{x}} \cdot \phi (\mathbf{x}) = \sum^d_{k=1} \partial_{x_k} \phi_k(\mathbf{x})$. It was established in \cite{ksd} that $\mathbb{S}(q, p) = 0 \iff p = q$ for a strictly positive-definite kernel $\mathcal{K}$.  To minimize the KL divergence, SVGD considers the incremental transform $\mathbf{x}_\epsilon = \mathbf{x} + \epsilon \phi(\mathbf{x})$, where $\mathbf{x} \sim q(\mathbf{x})$ and $\epsilon$ is a scalar step-size parameter. The resulting distribution after applying the transform is $q_{[\epsilon\phi]}$. SVGD uses the following result:
\begin{equation}
\nabla_{\epsilon} \mathcal{KL}(q_{[\epsilon\phi]}|| p(\mathbf{x}))|_{\epsilon=0} = -\mathbb{E}_{\mathbf x \sim q}[\mathcal{A}_p \phi(\mathbf x)],
\end{equation}
which relates the Stein operator and the derivative of the KL divergence w.r.t the perturbation $\epsilon$. We would like to select $\phi$ that maximally decreases the KL divergence. By considering  $\phi \in \{ \phi \in \mathcal{H}^d \: ; \: ||\phi||_{\mathcal{H}^d} \leq 1\}$, the optimal $\phi$ is the solution to the following constrained functional maximization:
\begin{equation}
\phi^* = \arg \max_{\phi \in \mathcal{H}^d} \{-\nabla_{\epsilon} \mathcal{KL}(q_{[\epsilon\phi]}|| p(\mathbf{x}))|_{\epsilon=0}, \text{s.t.} ||\phi||_{\mathcal{H}^d} \leq 1\}.
\label{eq:stein_problem}
\end{equation}
This maximization has a closed-form solution, derived by Liu et al. in Theorem 3.8 of \cite{ksd}. Note that we have used a slightly different definition of the Stein operator than that used by Liu et al., with $\mathcal{A}_p$ as defined in equation (\ref{eq:stein_op}) as equal to the trace of the Stein operator defined in \cite{ksd}. The closed-form solution is given by
\begin{align}
\phi^*(\cdot) &= \mathbb{E}_{\mathbf x \sim q}[\mathcal{A}_p \mathcal{K}(\cdot, \mathbf{x})] \\
 &= \mathbb{E}_{\mathbf x \sim q}[\mathcal{K} (\cdot, \mathbf x) \nabla_\mathbf{x} \log p(\mathbf x) + \nabla_\mathbf{x} \mathcal{K} (\cdot, \mathbf x)],
\label{eq:stein_soln}
\end{align}
and the resulting gradient of the KL divergence is
\begin{equation}
    \nabla_{\epsilon} \mathcal{KL}(q_{[\epsilon \phi^*]}|| p(\mathbf{x}))|_{\epsilon=0} = - \mathbb{S}(p, q).
\end{equation}
This implies that for a suitably-chosen kernel $\mathcal{K}$, if the gradient of the KL divergence is zero then the KSD is also zero, which means that $p = q$. We finally arrive at the update rule given in equation (\ref{eq:stein_update}) as the approximation of the closed-form solution in equation (\ref{eq:stein_soln}) with a finite set of particles.
\subsection{Orthogonal-Space Stein Variational Gradient Descent} \label{section:osvgd}
Recently Zhang et. al. proposed O-SVGD, a method for performing SVGD with a single equality constraint \cite{constrained_stein}, though they do not consider the problem of trajectory optimization. In this section, we give an overview of O-SVGD, but we give an alternative derivation to that given in \cite{constrained_stein} based on vector-valued RKHS and matrix-valued kernels \cite{matrix_kernel}. This alternative derivation will allow us to analyze our algorithm (Section \ref{methods:analysis}). The problem \cite{constrained_stein} aims to solve is
\begin{equation}
\min_q \mathcal{KL}(q(\mathbf{x}) || p(\mathbf{x})) \quad \text{s.t.} \quad  P_q(h(\mathbf{x}) = 0) = 1,
\end{equation}
where $h$ represents a single equality constraint. For particles $\mathbf{x}$ that are on the manifold induced by $h(\mathbf{x}) = 0$, we would like them to remain on the manifold after applying the Stein update in equation (\ref{eq:stein1}). To do this, we replace the function $\phi(\mathbf x)$ with $P(\mathbf x)\phi(\mathbf x )$. Where $P(\mathbf x)$ projects the updates to be in the tangent space of the constraint and is given by
\begin{align}
    P(\mathbf x) = I - \frac{\nabla h(\mathbf{x}) \nabla h(\mathbf{x})^T}{||\nabla(h(\mathbf{x})||^2}.
    \label{eq:ostein}
\end{align}
We can develop an SVGD algorithm that updates particles on the constraint manifold by considering the set of functions $\{P(\mathbf x)\phi(\mathbf x ), \; \phi(\mathbf{x}) \in \mathcal{H}^d\}$. By applying Lemma 2 from \cite{matrix_kernel} we establish that this set of functions is an RKHS $\mathcal{H}^d_\perp$ with matrix-valued kernel $\mathcal{K}_\perp$ given by
\begin{align}
    \mathcal{K}_\perp(\mathbf{x}^i, \mathbf{x}^j) &=  P(\mathbf{x}^i) \mathcal{K}(\mathbf{x}^i, \mathbf{x}^j) P(\mathbf{x}^j)^T \\
    &=  \mathcal{K}(\mathbf{x}^i, \mathbf{x}^j) P(\mathbf{x}^i) P(\mathbf{x}^j),
\end{align}
where we have used the fact that $\mathcal{K}$ is a scalar function and that $P(x)$ is symmetric to rearrange. Running SVGD with kernel $\mathcal{K}_\perp$ will therefore solve the constrained minimization problem (\ref{eq:stein_problem}), maximally reducing the KL divergence while only considering updates that lie in the tangent space of the constraint. 
Zhang et. al. \cite{constrained_stein} also add a term to equation (\ref{eq:stein1}) that drives particles to the manifold induced by the constraint
\begin{equation}
\phi_C = - \frac{\psi(h(\mathbf{x}))\nabla h(\mathbf{x})}{||\nabla h(x)||^2},
\label{eq:osvgd_phiC}
\end{equation}
where $\psi$ is an increasing odd function.

\section{Methods}
\label{sec:methods}

\begin{figure*}
\centering
\subfloat[]{\includegraphics[width=0.195\textwidth]{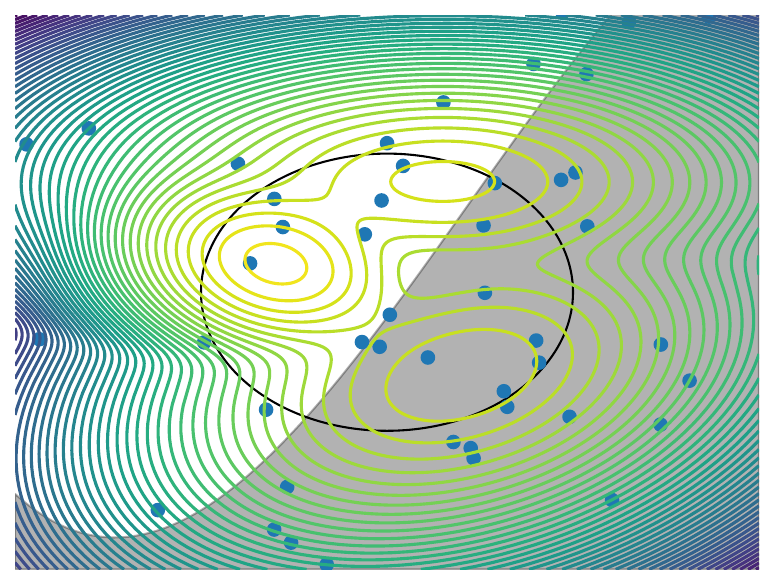}} \hfill 
\subfloat[]{\includegraphics[width=0.195\textwidth] {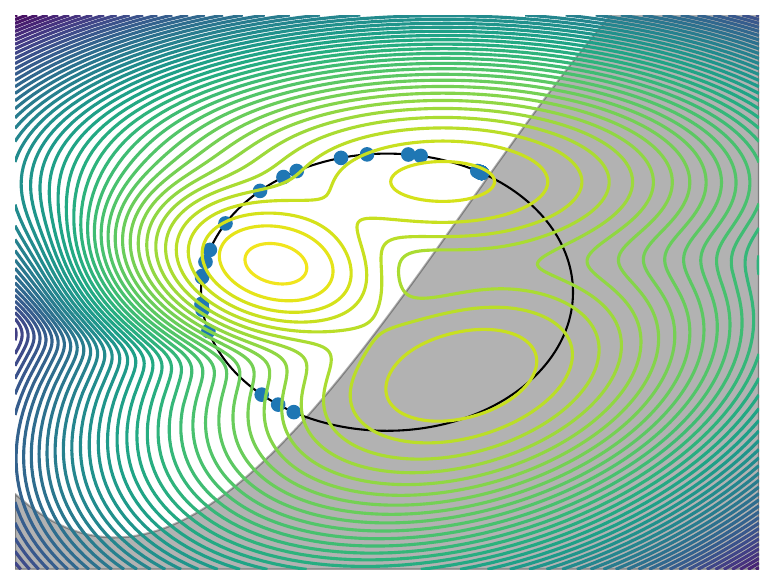}} \hfill 
\subfloat[]{\includegraphics[width=0.195\textwidth]{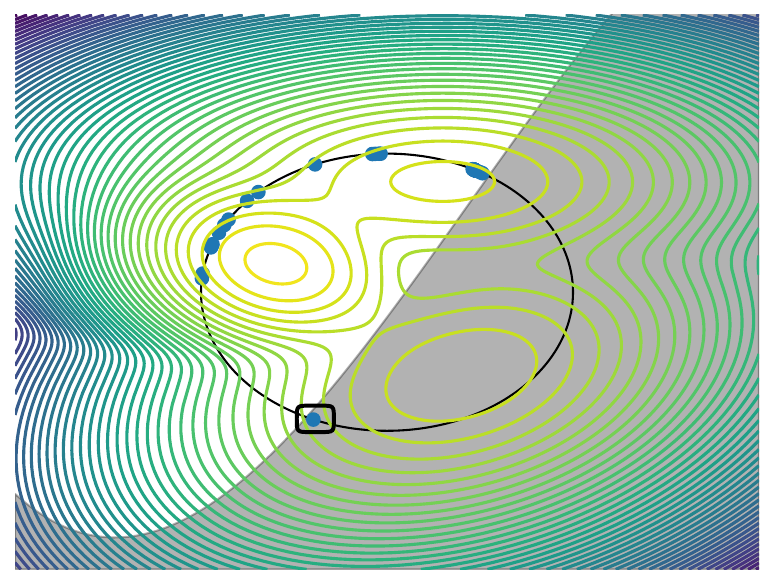}} \hfill 
\subfloat[]{\includegraphics[width=0.195\textwidth]{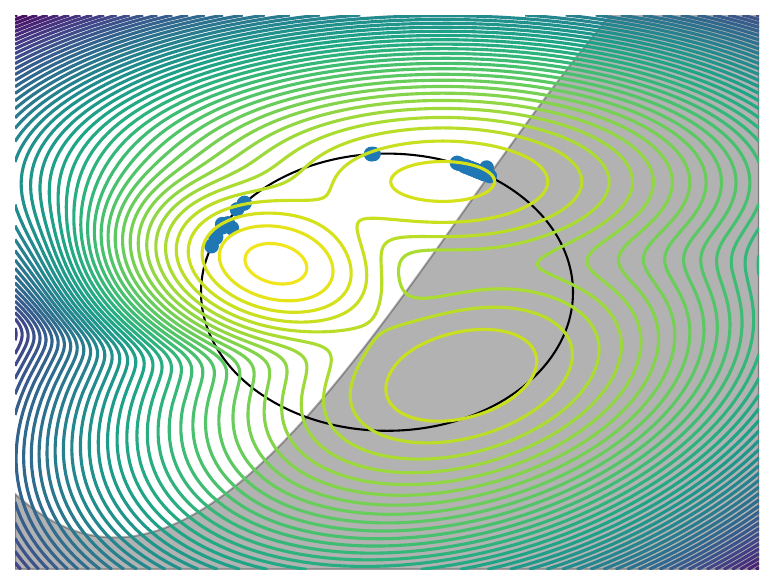}} \hfill 
\subfloat[]{\includegraphics[width=0.195\textwidth]{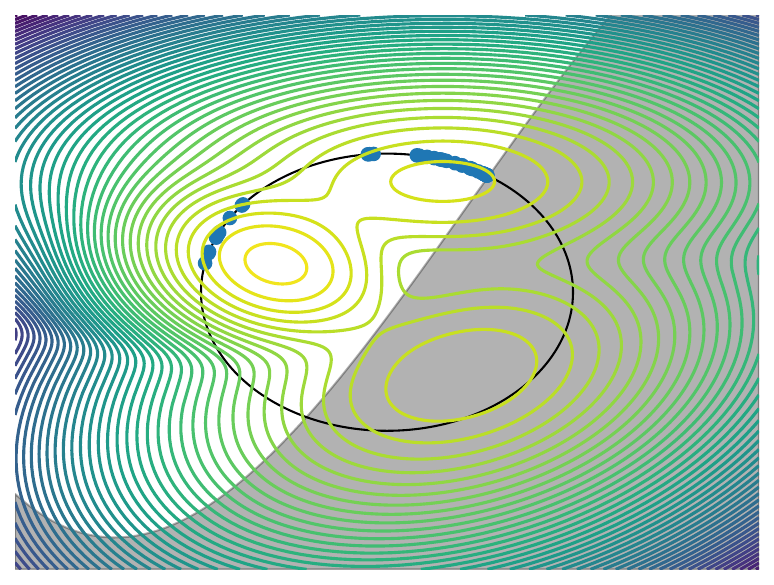}} \hfill 
\caption{CSVTO visualized for a 2D problem. The posterior is a mixture of 3 Gaussians, with the log posterior peaks visualized. There is an equality constraint that the particles must lie on the circle. There is also an inequality constraint that the particles must lie outside the shaded region. a) The initial particles are randomly generated and are not necessarily feasible. b) Due to the annealing discussed in section \ref{sec:methods:annealing}, early on in the optimization the particles are constraint-satisfying and diverse. c) The particles move towards the relative peaks of the objective, however, the circled particle has become stuck in a poor local minimum due to the constraints, where the gradient of the log posterior is directed towards an infeasible peak. Since the particle is isolated it is not sufficiently affected by the repulsive gradient term that would help escape the local minimum. d) The re-sampling step from section \ref{sec:methods:resampling} re-samples the particles, applying noise in the tangent space of the constraints. This eliminates the particle at the poor local minimum. e) The set of particles converges around the local minimum of the objectives while satisfying the constraints.}
\label{fig:method_toy}
\end{figure*}

Our proposed trajectory optimization algorithm uses SVGD to perform constrained optimization on a set of trajectories in parallel. The result is a diverse set of low-cost constraint-satisfying trajectories. The full algorithm is shown in Algorithm \ref{alg:csvgd}. First, we will introduce the main component of our proposed algorithm, which decomposes the Stein update into a step tangential to the constraint boundary, and a step toward constraint satisfaction. We will then provide an analysis of the algorithm which relates it to problem (\ref{eq:constrained_planning_inference}).  Finally, we will discuss strategies for improving performance which include separating the bounds constraints, an annealing strategy for increasing particle diversity, and re-sampling particles during the optimization. Figure \ref{fig:method_toy} demonstrates CSVTO being applied to a 2D toy problem. 

\subsection{Constrained Stein Trajectory Optimization}
\label{sec:methods:csvto}

Solving the constrained variational inference problem in (\ref{eq:constrained_planning_inference}) is very difficult, since it requires finding a distribution that may exhibit multi-modality and has constrained support. To address this, we use a non-parametric representation of the distribution $q(\tau)$. We use SVGD where each particle is a trajectory, and iteratively update the particle set while enforcing the constraints on each particle. To do this we extend O-SVGD to multiple equality and inequality constraints and use it to generate constraint-satisfying trajectories. 

First, we relate using SVGD for unconstrained trajectory optimization to the minimization of the unconstrained variational free energy $\mathcal{F}(q)$ from (\ref{background:free_energy}). Consider the iterative transform $\tau_\epsilon = \tau + \epsilon \phi^*(\tau)$, where $\phi^*$ is the solution to (\ref{eq:stein_problem}) with posterior log likelihood $\log p(\tau | o=1)$, $\tau \sim q(\tau)$ and $\tau_\epsilon \sim q_{[\epsilon \phi^*]}(\tau)$. We can recast (\ref{eq:stein_problem}) for trajectories in terms of the free energy $\mathcal{F}(q)$
\begin{equation}
\phi^*(\tau) = \arg \max_{\phi \in \mathcal{H}^d} \{-\nabla_{\epsilon} \mathcal{F}(q_{[\epsilon\phi]})|_{\epsilon=0}, \text{s.t.} ||\phi||_{\mathcal{H}^d} \leq 1\}.
\label{eq:stein_traj_problem}
\end{equation}
Thus the update $\tau + \epsilon \phi^*$ ensures we maximally decrease the variational free energy. If $\phi^*(\tau) = 0$ then $q(\tau)$ is at a local minimum of $\mathcal{F}(q)$. We will now modify the Stein update to account for constraints. 
\subsubsection{Equality constraints}
We propose a modified Stein update rule for the $i$-th particle, in which we decompose the update into two components:
\begin{equation}
    \mathbf \tau^i_{k+1} = \tau^i_k + \alpha_J \mathbf{\phi}_{\perp}(\tau^i_k) + \alpha_C \phi_C(\tau^i_k),
    \label{eq:csvto_update}
\end{equation}
where $\phi_\perp$ is an update that is tangential to the constraint boundary, $\phi_C$ acts in the direction that decreases constraint violation, $\alpha_J$ and $\alpha_C$ are scalar step size parameters, and $k$ is the iteration. We replace the O-SVGD $\phi_C$ from equation (\ref{eq:osvgd_phiC}) with a Gauss-Newton step to minimize $h(\tau)^T h(\tau)$
\begin{equation}
    \phi_C(\tau) = \nabla h(\tau)^T (\nabla h(\tau) \nabla h(\tau)^T)^{-1} \: h(\tau).
\end{equation}
This uses approximate second-order curvature information for fast convergence. We then compute the \textit{projection} matrix $P(\tau)$, which projects vectors onto the tangent space of the constraints as
\begin{equation}
    P(\tau) = I - \nabla h(\tau)^T (\nabla h(\tau) \nabla h(\tau)^T)^{-1} \nabla h(\tau).
    \label{eq:projection}
\end{equation}
Inverting $\nabla h(\tau) \nabla h(\tau)^T$ is only possible if $\nabla h(\tau)$ is full rank. While in Sections \ref{sec:methods:ineq} and \ref{methods:analysis} we assume that $\nabla h(\tau)$ is full rank, for numerical stability we compute the pseudo-inverse of $\nabla h(\tau) \nabla h(\tau)^T$ via the singular-value decomposition, discarding singular vectors corresponding to singular values that are smaller than $1\times10^{-6}$.
Once we have $P(\tau)$, we use it to define the tangent space kernel, as in \cite{constrained_stein}:
\begin{equation}
    \mathcal{K}_\perp(\tau^i, \tau^j) = \mathcal{K}(\tau^i, \tau^j) P(\tau^i) P(\tau^j).
\end{equation}
We then use this kernel for the SVGD update to produce an update that is in the tangent space of the constraint:
\begin{align}
\begin{split}
    \mathbf \phi^*_\perp(\tau^i) = &\frac{1}{N}\sum^N_{j=1} \mathcal{K}_\perp(\tau^i, \tau^j)   \nabla_{\tau^j} \log p(\tau^j |o=1)\\ &+ \nabla_{\tau^j} \mathcal{K}_\perp(\tau^i, \tau^j).
    \label{eq:total_update}
\end{split}
\end{align}
Since $\mathcal{K}_\perp$ is a matrix-valued function, the last term is calculated (as in \cite{matrix_kernel}) as
\begin{equation}
    [\nabla_{\tau^j} \mathcal{K}_\perp(\tau^i, \tau^j)]_l = \sum_m \nabla_{[\tau^j]_m} [\mathcal{K}_\perp(\tau^i, \tau^j)]_{l,m},
\end{equation}
where the notation $[x]_l$ indicates the $l$th element of $x$.
Equation (\ref{eq:total_update}) has several interesting features. First, two trajectory particles $\tau^i$ and $\tau^j$ are considered close if they are close according to the original kernel $\mathcal{K}$. In addition, expanding the first term to $\mathcal{K}(\tau^i, \tau^j)P(\tau^i)P(\tau^j) \nabla_{\tau^j}\log p(\tau^j|o=1)$, we see that if $P(\tau^i) = P(\tau^j)$ this reduces to $\mathcal{K}(\tau^i, \tau^j)P(\tau^j) \nabla_{\tau^j} \log p(\tau^j|o=1)$. For $P(\tau^i) \neq P(\tau^j)$, the magnitude of this term will always be reduced. Intuitively this means that particles will share gradients if particles are close and the tangent space of the constraint is similar. In addition, all updates will be in the tangent space of the constraint. \paragraph{Repulsive term in the tangent space} \label{sec:method:repulsion} The derivative $\nabla_{[\tau^j]_m} [\mathcal{K}_\perp(\tau^i, \tau^j)]_{l,m}$ can be expanded to
\begin{equation}
\begin{split}
    \nabla_{[\tau^j]_m} & [\mathcal{K}_\perp(\tau^i, \tau^j)]_{l,m} = \nabla_{[\tau^j]_m} [\mathcal{K}(\tau^i, \tau^j) P(\tau^i) P(\tau^j)]_{l,m} \\
    = \quad & [P(\tau^i) P(\tau^j)]_{l,m}\nabla_{[\tau^j]_m} \mathcal{K}(\tau^i, \tau^j) + \\ 
    & \mathcal{K}(\tau^i, \tau^j) [P(\tau^i)]_{l,m} \nabla_{[\tau^j]_m} [P(\tau^j)]_{l,m}.
\end{split}
\label{eq:grad_kernel}
\end{equation}
We see from equation (\ref{eq:grad_kernel}) above that the gradient of the kernel consists of two terms. The first term projects the gradient of the unconstrained kernel to the tangent space of the constraints both at $\tau^i$ and $\tau^j$.

The second term requires computing the derivative of the matrix-valued projection function. This term is expanded further in Appendix \ref{sec:appendix}, showing that it requires the evaluation of the second derivative of the constraint function $\nabla^2 h(\tau)$. For problems with constraints for which the second derivative is unavailable, we can remove this second term for individual constraints. We do this by setting the second derivative of a particular constraint to be the zero matrix (see Appendix \ref{sec:appendix}). Doing so effectively uses a locally linear approximation of the constraint to compute the repulsive gradient.

We will discuss how we define a kernel on trajectories in section \ref{methods:trajectory_kernel}. 
\subsubsection{Extension to Inequality Constraints}
\label{sec:methods:ineq}
We extend the above method to inequality constraints with the use of slack variables. We turn the inequality constraints into equality constraints with slack variable $\mathbf z$:
\begin{equation}
    g(\tau) + \frac{1}{2}\mathbf{z}^2 = 0.
\end{equation}
The full set of equality constraints then becomes
\begin{equation}
\hat{h} = \begin{bmatrix}
h(\tau) \\
g(\tau) + \frac{1}{2}\mathbf{z}^2
\end{bmatrix}.
\end{equation}
Converting inequality constraints to equality constraints via squared slack variables is often avoided as it can introduce spurious non-local-minima that satisfy the Karush–Kuhn–Tucker (KKT) conditions \cite{sq_slack}. To mitigate this issue we make an assumption on the regularity of the problem, denoted as (R) in \cite{ineq_flow}. The details of the assumption are technical and we do not include it here. 
The assumption essentially states that $\nabla \hat{h}$ is full rank at initialization and remains so during the optimization. Under these assumptions, Schropp \cite{ineq_flow} proved that the hyperbolic equilibrium points of the augmented system are local minima of the equality and inequality-constrained optimization problem. Optimizing multiple trajectories in parallel provides additional robustness against this issue. Even should some particles become stuck at one of these undesirable fixed points, in Section \ref{sec:methods:resampling} we propose a method for re-sampling the set of particles which redistributes particles away from these fixed points.
While we could avoid this issue by using non-negative slack variables with the transformation $g(\tau) + \mathbf{z}$, where $\mathbf{z} > 0$, we found that this led to poorer constraint satisfaction in practice.

After introducing the slack variables, we compute the constrained Stein update with all constraints as equality constraints. 
We augment the state with $\mathbf{z}$ as 
\begin{equation}
    \hat{\tau} = \begin{bmatrix}
        \tau \\
        \mathbf{z}
    \end{bmatrix}.
\end{equation}
The projection is given by
\begin{equation}
    P(\mathbf{\hat{\tau}}) = I - \nabla \hat{h}(\hat{\tau})^T (\nabla \hat{h}(\hat{\tau}) \nabla \hat{h}(\hat{\tau})^T)^{-1}\nabla \hat{h}(\hat{\tau}),
\end{equation}
and the kernel is
\begin{equation}
\mathcal{K}_\perp(\mathbf{\hat{\tau}}^i, \mathbf{\hat{\tau}}^j) = \mathcal{K}(\tau^i, \tau^j) P(\mathbf{\hat{\tau}}^i) P(\mathbf{\hat{\tau}}^j).
\end{equation}
Notice that the kernel uses the original $\tau$ and not the augmented $\hat{\tau}$. We then perform the constrained Stein update on the augmented state:
\begin{equation}
    \begin{split}
        \mathbf \phi^*_\perp(\hat{\tau}^i) = &\frac{1}{N}\sum^N_{j=1} \mathcal{K}_\perp(\hat{\tau}^i, \hat{\tau}^j) 
        \begin{bmatrix}\nabla \log p(\tau^j | o=1) \\
        0
        \end{bmatrix} \\
        & + \nabla_{\hat{\tau}^j} \mathcal{K}_\perp(\hat{\tau}^i, \hat{\tau}^j) 
    \end{split}
    \label{eq:phi_j}
\end{equation}
\begin{equation}
    \phi_C(\hat{\tau}) = \nabla \hat{h}(\hat{\tau})^T (\nabla \hat{h}(\hat{\tau}) \nabla \hat{h}(\hat{\tau})^T)^{-1} \: \hat{h}(\hat{\tau}).
    \label{eq:phi_c}
\end{equation}

Once we have performed the iterative optimization we have a set of trajectories. We then select a trajectory to execute by choosing the one that minimizes the penalty function
\begin{equation}
\hat{C}_\lambda(\hat{\tau}) = C(\tau) + \lambda \sum |\hat{h}(\hat{\tau})|.
\end{equation}
\subsubsection{Analysis}
\label{methods:analysis}
In this section, we provide an analysis of CSVTO. We demonstrate that stationary points of the gradient flow satisfy the first-order optimality conditions for the constrained variational optimization problem in (\ref{eq:constrained_planning_inference}), subject only to equality constraints. 
\begin{thm}\label{thm1}
Assume that $\nabla h$ is full rank. Let $\phi^* \in \mathcal{H}^d$ be the solution to (\ref{eq:stein_problem}) with the unconstrained kernel $\mathcal{K}$, and $\phi^*_\perp \in \mathcal{H}^d_\perp$ be the solution to (\ref{eq:stein_problem}) using the tangent space kernel $\mathcal{K}_\perp$. If the following holds:
\begin{equation}
    \alpha_J \phi^*_\perp(\tau) + \alpha_C \phi_C(\tau) = 0,
    \label{stein_stationary}
\end{equation}
then the following must be true:
\begin{align}
    \phi^*(\tau) + \nabla h(\tau)^T \mu = 0 \label{stationarity}\\
    h(\tau) = 0,
    \label{feasibility}
\end{align}
where $\mu$ is a vector of Lagrange multipliers. 
\begin{proof}
Since $\phi_C$ and $\phi^*_\perp$ are orthogonal, then if equation (\ref{stein_stationary}) holds then $\phi_C = \phi^*_\perp = 0$. Next, we note that $\phi^*_\perp(\tau) = P(\tau)\hat{\phi}$, where $\hat{\phi} \in \mathcal{H}^d$ and further $P(\tau)\hat{\phi}(\tau) = 0 \implies P(\tau)\phi^*(\tau) = 0$. To see this, consider $P(\tau)\phi^*(\tau) \neq 0$. This would imply that $\nabla_{\epsilon} \mathcal{KL}(q_{[\epsilon P\phi^*]}|| p(\tau|o=1))|_{\epsilon=0} \neq 0$, which implies that there is a descent direction. This would mean that $\exists \: \phi_{\perp}$ such that $-\nabla_{\epsilon} \mathcal{KL}(q_{[\epsilon \phi_\perp^*]}|| p(\tau|o=1))|_{\epsilon=0} < -\nabla_{\epsilon} \mathcal{KL}(q_{[\epsilon \phi_\perp}||p(\tau|o=1))_{\epsilon=0}$, which is a contradiction. Expanding $P(\tau) \phi^* = 0$ yields
\begin{equation}
\begin{split}
    \left [ I - \nabla h(\tau)^T \left( \nabla h(\tau) \nabla h(\tau)^T \right) ^{-1} \nabla h(\tau) \right ] \phi^*(\tau) = 0 \label{stationarity_phiJ} \\
    \phi^*(\tau) - \nabla h(\tau)^T \left [ \left(\nabla h(\tau) \nabla h(\tau)^T \right )^{-1} \nabla h(\tau) \phi^*(\tau) \right ] = 0.
\end{split}
\end{equation}
Specifying $\mu = -\left(\nabla h(\tau) \nabla h(\tau)^T\right)^{-1} \nabla h(\tau) \phi^*(\tau)$ results in equation (\ref{stationarity}) being satisfied. Now we expand $\phi_C = 0$ resulting in
\begin{equation}
    \nabla h(\tau)^T (\nabla h(\tau) \nabla h(\tau)^T)^{-1} h(\tau) = 0. \label{stationary_phic}
\end{equation}
To show feasibility at the stationary point we left multiply (\ref{stationary_phic}) by $\nabla h(\tau)$, which for full rank $\nabla h$ results in $h(\tau) = 0$.  
\end{proof}
\end{thm}
Theorem \ref{thm1} holds when we can integrate the expectation in (\ref{eq:stein_soln}). However, we are approximating the expectation with particles so (\ref{stationarity}) may not hold in practice. However, the feasibility condition (\ref{feasibility}) remains true when using a particle approximation for $q$.
To extend this proof to inequality constraints, note that in Section \ref{sec:methods:ineq} we discussed the regularity conditions under which hyperbolic stable stationary points of the gradient flow on the augmented equality-constrained system satisfy first-order optimality conditions of the original system with both equality and inequality constraints.
\subsubsection{Annealed SVGD for improved diversity}
\label{sec:methods:annealing}
We employ an annealing technique for SVGD as proposed in \cite{annealed_svgd}. We use a parameter $\gamma \in [0, 1]$ which controls the trade-off between the gradient of the posterior log-likelihood and the repulsive gradient. For $\gamma << 1$ the repulsive term dominates resulting in trajectories being strongly forced away from one another. As $\gamma$ increases the gradient of the posterior likelihood has a greater effect resulting in trajectories being optimized to decrease the cost. When combined with $\phi_C$ this results in the optimization prioritizing diverse constraint-satisfying trajectories first, then decreasing cost later in the optimization. The annealed update is given by  
\begin{equation}
    \begin{split}
        \mathbf \phi^i_\perp(\hat{\mathbf{\tau}}) = &\frac{1}{N}\sum^N_{j=1} \gamma \mathcal{K}_\perp(\mathbf{\hat{\tau}}_i, \mathbf{\hat{\tau}}_j) 
        \begin{bmatrix}\nabla \log p(\tau_j|o) \\
        0
        \end{bmatrix} \\
        & + \nabla_{\mathbf{\hat{\tau}}_j} \mathcal{K}_\perp(\mathbf{\hat{\tau}}_i, \mathbf{\hat{\tau}}_j). 
    \end{split}
    \label{eq:stein_anneal}
\end{equation}
We use a linear annealing schedule with $\gamma_k = \frac{k}{K}$, where $K$ is the total number of iterations. When performing online re-planning, we only perform the annealing when optimizing the trajectory the first time-step. 
\subsubsection{Trajectory Kernel}
\label{methods:trajectory_kernel}
CSVTO relies on a base kernel $\mathcal{K}(\tau^i, \tau^j)$ operating on pairs of trajectories which defines the similarity between trajectories. As noted by Lambert et. al. \cite{stein_mpc}, high dimensional spaces can result in diminishing repulsive forces, which can be problematic for trajectory optimization problems due to the time horizon. We use a similar approach to SVMPC \cite{stein_mpc} in that we decompose the kernel into the sum of kernels operating on smaller components of the trajectory. We use a sliding window approach to decompose the trajectory. For a given sliding window length $W$ let $\tau^t = [x_{t:t+W}, u_{t-1:t-1+W}]^T$. The overall kernel is then given by
\begin{equation}
    \mathcal{K}(\tau^i, \tau^j) = \frac{1}{T-W}\sum_t^{T-W} \mathcal{K}(\tau_t^i, \tau_t^j).
\end{equation}
We use the Radial-Basis Function (RBF) kernel $\mathcal{K}(\tau_t^i, \tau_t^j) = \exp (-\frac{1}{h}||\tau^i_t - \tau^j_t||_2^2)$ as the base kernel, where $h$ is the kernel bandwidth. We use the median heuristic as in \cite{liu2016stein} to select the kernel bandwidth:
\begin{equation}
    h = \frac{\texttt{median}(||\tau^i_t - \tau^j_t||_2)^2}{\log (N)},
\end{equation}
where $N$ is the number of particles. 
\subsubsection{Bounds constraint}
Bounds constraints can, in principle, be handled as general inequality constraints as described in the above section. However, since this involves adding additional slack variables incorporating bounds constraints involves an additional $T \times 2(d_u + d_x)$ decision variables in the optimization, where $d_x$ and $d_u$ are the state and control dimensionalities, respectively. It is more computationally convenient to use a simple approach where at every iteration we directly project the trajectory to satisfy the bound constraints. This is done by
\begin{align}
    \tau^* &= \min(\max(\tau_{min}, \tau), \tau_{max}).
\end{align}
\subsubsection{Initialization} As introduced in section \ref{sec:vi_trajopt}, we have a user-specified prior on controls $p(U)$. To initialize CSVTO on a new problem, we proceed by sampling from this prior $p(U)$ and using the dynamics $f(x_t, u_t)$ to generate sampled trajectories. In this way, we ensure that the initial trajectory satisfies the dynamics constraints.

We use a different initialization scheme when running trajectory optimization online in a receding horizon fashion as in Algorithm \ref{alg:csvto_online}, as it is typical to warm-start the optimization with the solution from the previous timestep. For a single particle, the trajectory consists of $\tau = (\mathbf{x}_1, ..., \mathbf{x}_T, \mathbf{u}_0, ..., \mathbf{u}_{T-1)})$. The shift operation computes $\tau' = (\mathbf{x}_2, ..., \mathbf{x}'_{T+1}, \mathbf{u}_1, ..., \mathbf{u}'_T)$. Here $\mathbf{x}'_{T+1}, \mathbf{u}'_T$ is the initialization for the newly considered future timestep. The initializations $\mathbf{x}'_{T+1}, \mathbf{u}'_T$ may be chosen in a problem-specific way. In our approach, we choose them by duplicating the previous timestep's state and control, i.e. $(\mathbf{x}'_{T+1}, \mathbf{u}'_T) = (\mathbf{x}_T, \mathbf{u}_{T-1})$.

When running the algorithm with inequality constraints, for both the online and warm-start optimizations we initialize the slack variable $\mathbf{z}$ with $\mathbf{z} = \sqrt{2 |g(\tau)|}$ so that trajectories satisfying the inequality constraint are initialized to satisfy the transformed equality constraint.

The above heuristic is motivated by the assumption that the solution should not vary much between timesteps. However, the fact that we have a set of trajectories rather than a single one can invalidate this assumption, since we can only take a single action. Trajectories that have very different first actions from the action taken can end up being quite poor initializations, particularly in the presence of constraints that can render them infeasible. Over time these poor initializations can lead to the degradation in the quality of the particles, which motivates the next section in which we discuss a re-sampling technique to prevent sample impoverishment. 

\subsubsection{Re-sampling}
\label{sec:methods:resampling}
As discussed above, the shift operation can lead to trajectories that are not executed becoming infeasible and rendering those particles useless for trajectory optimization. In addition, our cost and constraints are not necessarily convex, so, as with any local optimization method, poor initializations can lead to infeasibility. We take inspiration from the Particle Filter literature \cite{particle_filter_resampling} and incorporate a re-sampling step which is executed when performing online re-planning. Every \texttt{resample\_steps} timesteps we re-sample after performing the shift operation on the previous trajectory particles. To perform re-sampling, we compute weights using the penalty function
\begin{equation}
w_i = \frac{\exp(-\frac{\hat{C}_\lambda(\hat{\tau}_i)}{\beta})}{\sum^N_j\exp(-\frac{\hat{C}_\lambda(\hat{\tau}_j)}{\beta})},
\end{equation}
where $\beta$ is a temperature parameter. We then re-sample a new set of particles according to weights $w_i$. It is common in the particle filter literature to additionally add noise, to prevent re-sampled particles collapsing. However, in our case, it is undesirable to add random noise to a constraint-satisfying trajectory as it may lead to constraint violation. 
We avoid this issue by sampling noise and projecting the noise to only have components in the tangent space of the constraints for a given trajectory. Suppose we have sampled trajectory $\tau_i$ from the set of particles. We first sample $\epsilon \sim \mathcal{N}(0, \sigma_{resample}^2 I)$, and then update the trajectory with
\begin{equation}
    \tau_{new} = \tau_i + P(\tau_i)\epsilon,
\end{equation}
where $P(\tau_i)$ is the projection matrix from equation (\ref{eq:projection}).

\begin{algorithm}
\caption{A single step of CSVTO, this will run every timestep.}
\begin{algorithmic}[1]
    \Function{CSVTO}{$\mathbf{x}_0, \tau, K, \texttt{anneal}$}
    \State $\mathbf{z} \gets \sqrt{2 |g(\tau)|}$
    \State $\hat{\tau} \gets [\tau, \mathbf{z}]^T$
    \For{k $\in \{1, ..., \text{K}$\}}
        \For{i $\in \{i, ..., \text{N}$\}}
            \State$\phi^i_C \gets$ via eq. \ref{eq:phi_c}
            \If {\texttt{anneal}}
                \State $\gamma \gets \frac{k}{K}$
                \State$\phi^i_\perp \gets$ via eq. \ref{eq:stein_anneal}
            \Else 
                \State$\phi^i_\perp \gets$ via eq. \ref{eq:phi_j}
            \EndIf
            \State $\hat{\tau}^i \gets \hat{\tau}^i + \alpha_J \mathbf{\phi}^i_{\perp} + \alpha_C \phi^i_C$
            \State $\hat{\tau}^i \gets $ \textproc{ProjectInBounds}$(\hat{\tau}^i)$. 
        \EndFor
    \EndFor
    \LineComment{Get the best trajectory according to penalty function}
    \State $\hat{\tau}^* \gets \arg \min_\tau \hat{C}_\lambda(\hat{\tau})$
    \LineComment{Discard slack variables}
    \State $\tau^*, \tau \gets \hat{\tau}^*, \hat{\tau}$
    \State \Return $\tau^*, \tau$
\EndFunction
\end{algorithmic}
\label{alg:csvgd}
\end{algorithm}

\begin{algorithm}
\caption{CSVTO running with online re-planning}
\begin{algorithmic}[1]
    \Function{CSVTO\_MPC}{$\mathbf{x}_0, \tau_0$}
    \For{t $\in \{1, ..., \text{T}$\}}
        \LineComment{Resample}   
        \If{\textproc{mod}(t, \texttt{resample\_steps}) $= 0$}
            \State $\tau_t \gets$ \textproc{Resample}($\tau_t$)
        \EndIf
        \If {t = 1}
            \State $K \gets K_{w}$
            \State $\texttt{anneal} \gets \texttt{True}$
        \Else
            \State $K \gets K_{o}$
            \State $\texttt{anneal} \gets \texttt{False}$
        \EndIf
        \State $\tau^*_{t}, \tau_t \gets$ \textproc{CSVTO}$(x_t, \tau_t, K, \texttt{anneal})$ 
        
        \LineComment{Select first control from the best trajectory}
        \State $\mathbf{u}_t \gets \tau*$
        \State $\mathbf{x}_t \gets$ \textproc{StepEnv}$(\mathbf{u}_t)$
        \LineComment{Shift operation}
        \State $\tau_{t+1} \gets$ \textproc{Shift}($\tau_t$)
        
    \EndFor
    
\EndFunction
\end{algorithmic}
\label{alg:csvto_online}
\end{algorithm}
\section{Evaluation}
\label{sec:eval}

We evaluate our approach in three experiments. The first is a constrained 12DoF quadrotor task which has nonlinear underactuated dynamics. The second experiment is a 7DoF robot manipulator task, where the aim is to move the robot end-effector to a goal location while being constrained to move along the surface of a table. The third experiment is also a 7DoF robot manipulator task, where the aim is to manipulate a wrench to a goal angle. Both of these 7DoF manipulator tasks involve planning in highly constrained domains. We perform the manipulator experiments in IsaacGym \cite{isaacgym}. The hyperparameters we use in all experiments are shown in Table \ref{table:hyperparameters}. For all experiments, the costs and constraint functions are written using PyTorch \cite{pytorch}, and automatic differentiation is used to evaluate all relevant first and second derivatives. 
\subsection{Baselines}
We compare our trajectory optimization approach to both sampling-based and gradient-based methods. We compare against IPOPT \cite{ipopt}, a general non-linear constrained optimization solver, which has been widely used for robot trajectory optimization \cite{ipopt_trajectory1, ipopt_trajectory2}. We use the MUMPS \cite{mumps} linear solver for IPOPT. When running IPOPT, where second derivatives are available we use exact derivatives computed via automatic differentiation in PyTorch, where they are not available we use the Limited-memory Broyden–Fletcher–Goldfarb–Shanno algorithm (L-BFGS) \cite{liu1989limited} to approximate the Hessian. The method used will be indicated for each experiment. For CSVTO  and IPOPT, we use a direct transcription scheme; IPOPT solves the optimization problem as expressed in problem (\ref{eq:trajectory_opt}). For IPOPT we use the open-source implementation provided by \cite{ipopt}.

We additionally compare against MPPI \cite{mppi} and SVMPC \cite{stein_mpc}. MPPI and SVMPC are methods for performing unconstrained trajectory optimization, with constraints commonly incorporated with penalties. For these methods we use the penalty function $\hat{C}_{(\lambda, \mu)}(\tau) = C(\tau) + \lambda \sum |h(\tau)| + \mu \sum |g(\tau)|^+$, where $|g(\tau)|^+$ is a vector consisting of only the positive values of $g(\tau)$. We use separate penalty weights for equality and inequality constraints. We evaluate each of these baselines on two different magnitudes of penalty weights on equality constraints $\lambda$. In the SVMPC paper, the authors show that their method can be used both with and without gradients. We evaluate against two versions of SVMPC, one using a sample-based approximation to the gradient and another using the true gradient. For SVMPC and MPPI we use a shooting scheme since they can only handle constraints via penalties, which can lead to poor satisfaction of the dynamics constraint. We use our own implementations for MPPI and SVMPC in PyTorch. 

\subsection{12DoF Quadrotor}
\label{sec:eval_quadrotor}
\begin{figure*}
\begin{tabular}{cccc}
\subfloat[CSVTO t=1]{\includegraphics[width=0.24\textwidth, trim=50 50 50 50,clip]{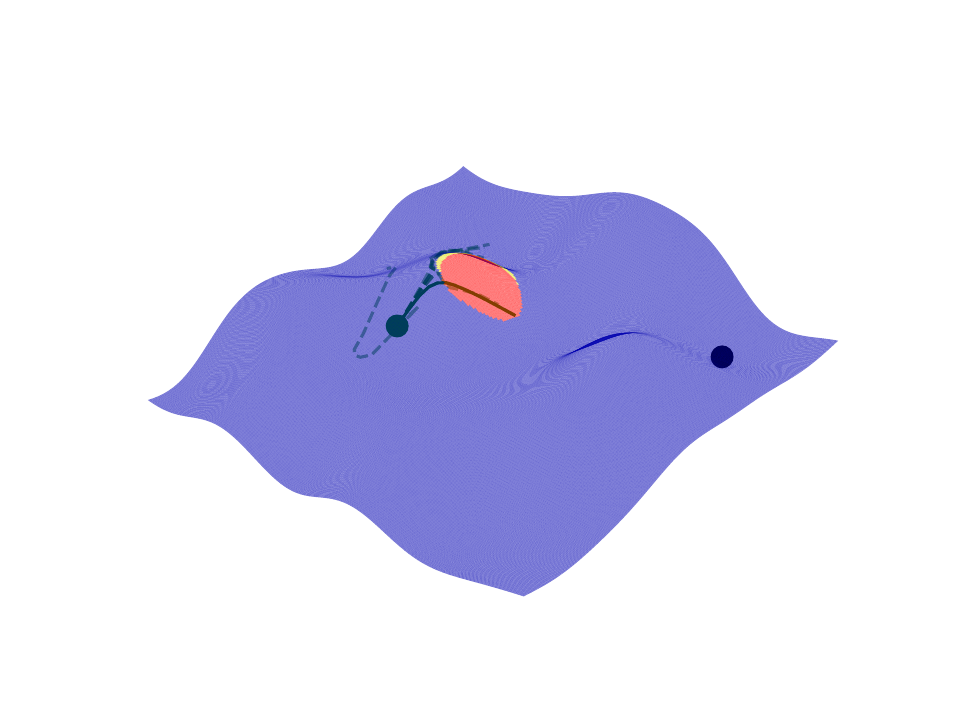}} &
\subfloat[CSVTO t=5]{\includegraphics[width=0.24\textwidth,trim=50 50 50 50,clip]{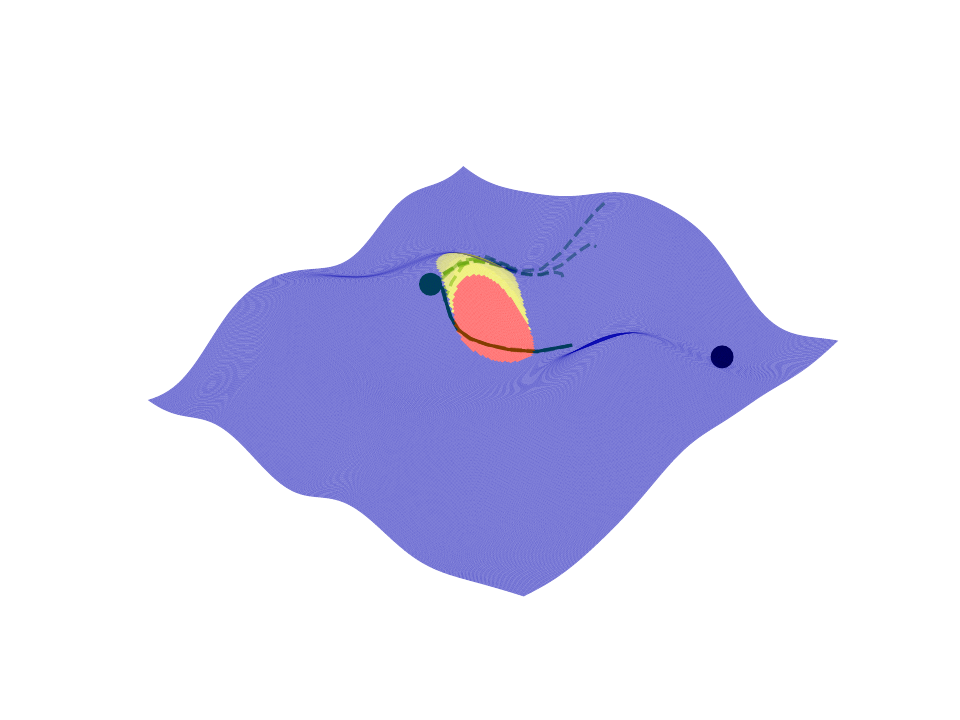}} &
\subfloat[CSVTO t=10]{\includegraphics[width=0.24\textwidth,trim=50 50 50 50,clip]{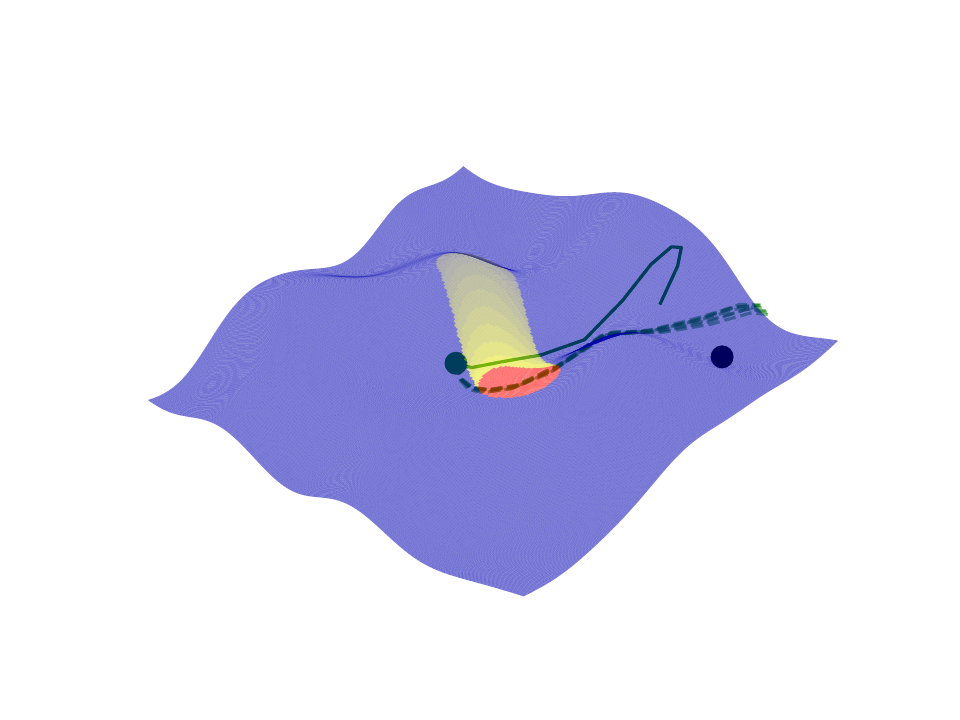}} &
\subfloat[CSVTO t=15]{\includegraphics[width=0.24\textwidth,trim=50 50 50 50,clip]{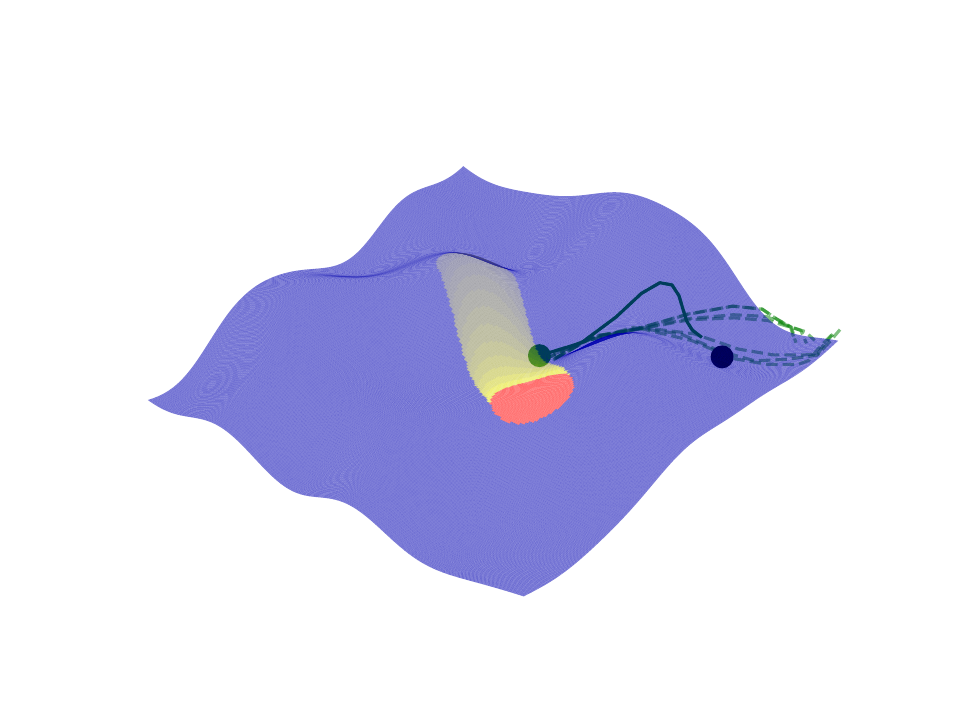}}\\
\subfloat[IPOPT t=1]{\includegraphics[width=0.24\textwidth,trim=50 50 50 50,clip]{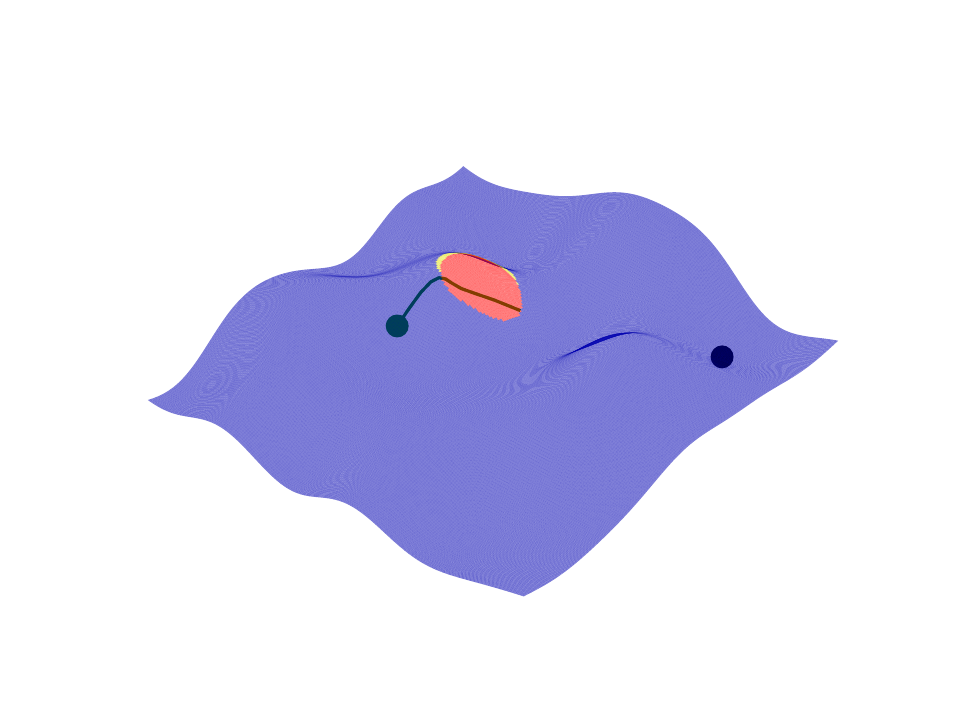}} &
\subfloat[IPOPT t=5]{\includegraphics[width=0.24\textwidth,trim=50 50 50 50,clip]{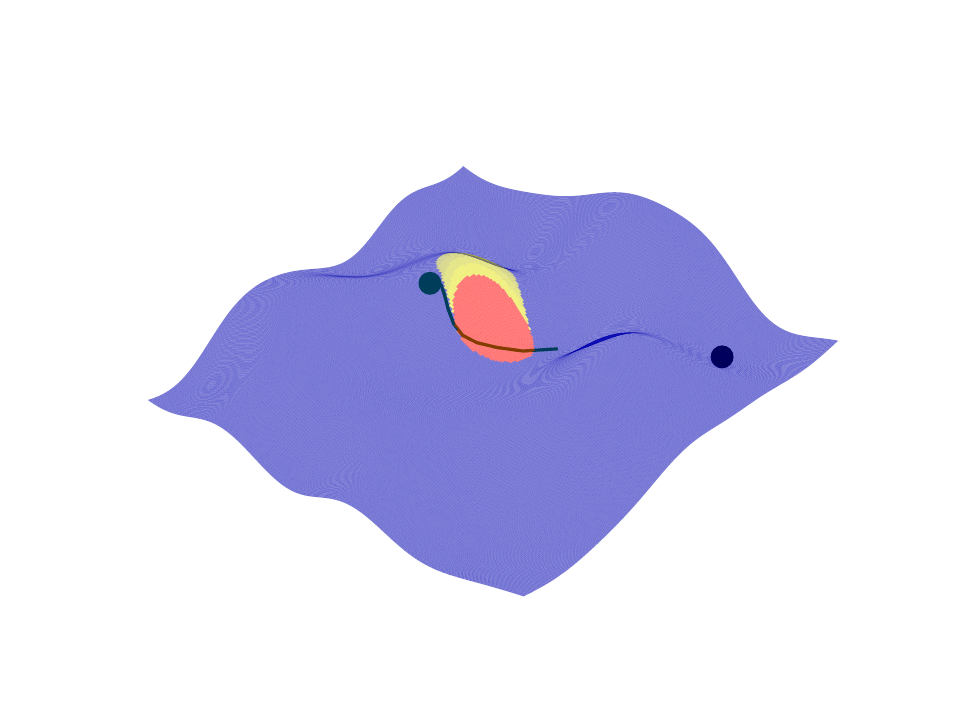}} &
\subfloat[IPOPT t=10]{\includegraphics[width=0.24\textwidth,trim=50 50 50 50,clip]{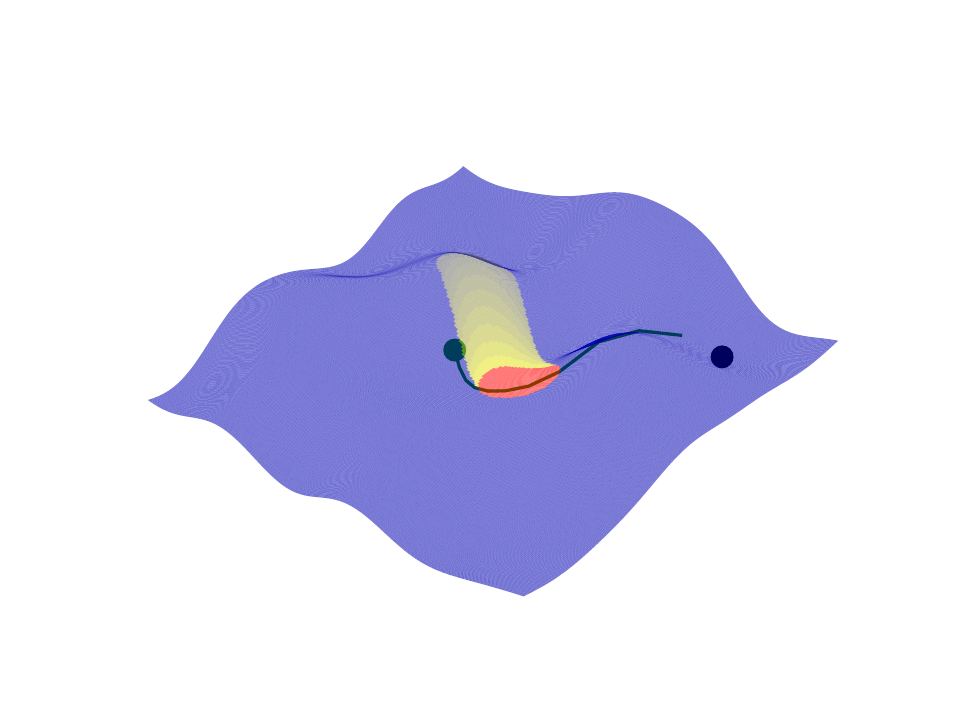}} &
\subfloat[IPOPT t=15]{\includegraphics[width=0.24\textwidth,trim=50 50 50 50,clip]{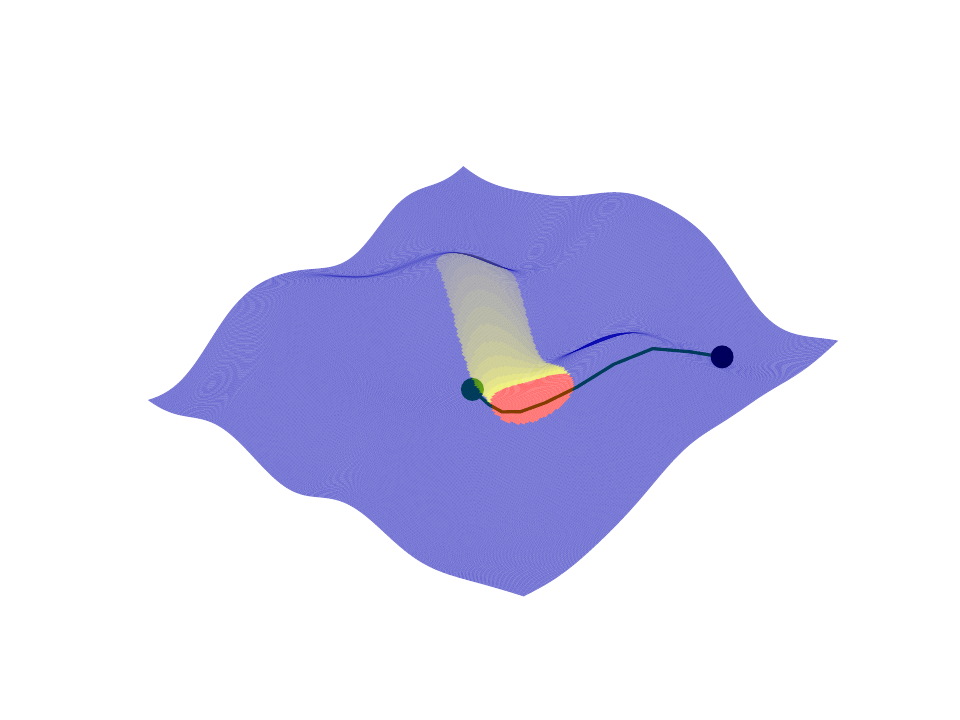}}
\end{tabular}
\caption{Experimental setup for the quadrotor task. The quadrotor must travel to the goal location, avoiding the obstacle in red while remaining on the blue manifold. The fading yellow shows the path of the obstacle from previous timesteps. a-d) CSVTO maintains a set of trajectories (dashed), with the selected trajectory shown as a solid curve. CSVTO can keep a diverse set of trajectories and switches between them to avoid the moving obstacle. e-f) IPOPT generates an initial trajectory that makes good progress toward the goal and obeys the manifold constraint. However, even after the first timestep the obstacle has moved to render this trajectory infeasible. As the obstacle moves further IPOPT is unable to find an alternative trajectory and ends in a collision.}
\label{fig:quadrotor_exp}
\end{figure*}

\begin{figure}
    \centering
    \includegraphics[width=0.3\textwidth]{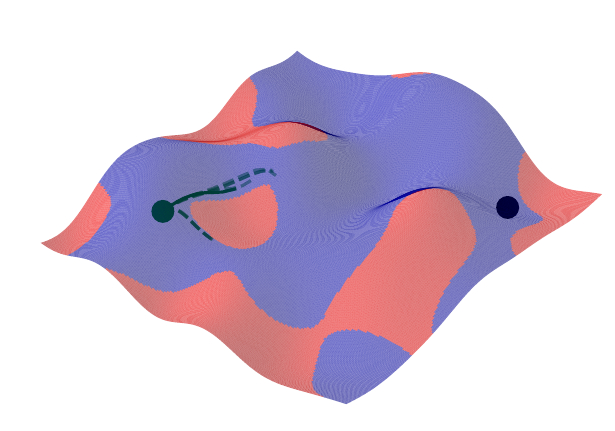}
    \caption{Experimental set-up for the quadrotor with static obstacles task. The quadrotor must travel to the goal location, avoiding the obstacles in red while remaining on the blue manifold.}
    \label{fig:quadrotor_static}
\end{figure}

We evaluate our method on a 12DoF underactuated quadrotor problem. The goal is to navigate the quadrotor from a start state to a goal state. We chose this problem to demonstrate our approach on a problem with complex nonlinear underactuated dynamics. The experimental setup is shown in Figures \ref{fig:quadrotor_exp} and \ref{fig:quadrotor_static}. The state of the quadrotor is $\mathbf{x} = [x, y, z, p, q, r, \dot{x}, \dot{y}, \dot{z}, \dot{p}, \dot{q}, \dot{r}]^T$, where $(x, y, z)$ is the 3D position and $(p, q, r)$ are the Euler angles. The control is the thrust $\mathbf{u} = [u_1, u_2, u_3, u_4]^T \in \mathbb{R}^4$. We place bounds constraints on the $(x, y)$ location of the quadrotor to be within a $10m \times 10m$ area centred at $(0, 0)$. The goal is to travel from start locations sampled uniformly from $x, y \in [3.0m, 4.5m]$ to a goal location of $(4, 4)$ within 100 time steps. We place an equality constraint that the quadrotor must travel along a nonlinear surface $z = f_{surf}(x, y)$. For this surface, we sample $z$ values from a Gaussian Process (GP) prior with an RBF kernel and zero mean function on a $10 \times 10$ grid of $(x, y)$ points. We use the sampled values as observations for a GP with the same kernel and mean function and use the corresponding posterior mean function as the equality constraint. We sample a single surface in this way and use it for all experiments. The dynamics for the 12DoF quadrotor are from \cite{quadrotor_dynamics} and are given by
\begin{equation}
    \begin{bmatrix}
    x \\
    y \\
    z \\
    p \\
    q \\
    r \\
    \dot{x} \\
    \dot{y} \\
    \dot{z} \\
    \dot{p} \\
    \dot{q} \\
    \dot{r} \\
    \end{bmatrix}_{t+1} = 
    \begin{bmatrix}
    x \\
    y \\
    z \\
    p \\
    q \\
    r \\
    \dot{x} \\
    \dot{y} \\
    \dot{z} \\
    \dot{p} \\
    \dot{q} \\
    \dot{r} \\
    \end{bmatrix}_t + \Delta t
    \begin{bmatrix}
    \dot{x} \\
    \dot{y} \\
    \dot{z} \\
    \dot{p} + \dot{q} s(p) t(q) + \dot{r} c(p) t(q) \\
    \dot{q} c(p)  - \dot{r} s \dot{p} \\
    \dot{q} \dfrac{s(p)}{c(q)} + \dot{r} \dfrac{c(p))}{c(q)} \\
    - ( s (p) s (r) + c(r) c(p) s(q)) K \frac{u_1}{m} \\
    - (c(r) s(p) - c(p) s(r) s (q)) K \frac{u_1}{m} \\
    g - c(p) s(q)) K \frac{u_1}{m} \\
    \frac{(I_y - I_z) \dot{q} \dot{r} + K u_2}{I_x} \\
    \frac{(I_z - I_x) \dot{p} \dot{r} + K u_3}{I_y} \\
    \frac{(I_x - I_y) \dot{p} \dot{q} + K u_4}{I_z} \\
    \end{bmatrix}_t 
\end{equation}
where $c(p), s(p), t(p)$ are $\cos, \sin, \tan$ functions, respectively. We use parameters $m=1 kg, I_x=0.5 kg\cdot m^2, I_y=0.1 kg\cdot m^2, I_z=0.3 kg\cdot m^2, K=5, g=-9.81 m\cdot s^{-2}$. We use the same dynamics both for planning and for simulation. 

We consider three variants of this task with different obstacle avoidance constraints: 1) We consider the case with no obstacles; 2) We consider the case of static obstacles. For the static obstacles case, we wish to demonstrate our method in a cluttered environment with arbitrarily shaped obstacles. We do this by generating the obstacles similarly to the surface constraint, which results in a smooth obstacle constraint. We consider a constraint function $f_{obs}(x, y)$, where the obstacle-free region is $\{(x, y), f_{obs}(x, y) \leq 0 \}$. We sample values for $f_{obs}(x, y)$ from a GP prior with an RBF kernel and a constant mean function of $-0.5$, so that there is a bias towards being obstacle-free, on a $10 \times 10$ grid of $(x, y)$ points. We then use these points as the observations for a GP with the same mean function and kernel as the GP prior. We also add observations at $(-4, -4)$ and $(4, 4)$ of $-2$, to ensure the start and goal regions are obstacle free. We use the resulting GP posterior mean function as $f_{obs}(x, y)$. We do this once and keep the same obstacle constraint for all trials. The resulting obstacle constraint is shown in Figure \ref{fig:quadrotor_static};
3) Finally, we consider a cylindrical obstacle in the x-y plane that moves during the trial in a path that is unknown to the planner; at every timestep, the planner plans assuming the obstacle will remain fixed. If the quadrotor collides with an obstacle during execution then we consider the task failed.

The planning horizon is 12. The posterior $\log p(\tau | o)$ for this problem is a quadratic cost given by 
\begin{equation}
    \begin{split}
    \log p(\tau | o) &= (x_T - x_{goal})^TP(x_T - x_{goal}) + \\ 
    & \sum^{T-1}_{t=1} (x_t - x_{goal})^T Q(x_t - x_{goal}) + u_{t-1}^T R u_{t-1}.
    \end{split}
\end{equation}
The control cost is equivalent to the prior on controls $p(u_t) = \mathcal{N}(0, 2R^{-1})$. The values we use for the costs are 
\begin{align}
Q &= \text{Diag}(5, 5, 0.5, 2.5, 2.5, 0.025, 1.25, 1.25, 1.25, 2.5, 2.5, 2.5) \\
P &= 2Q\\ 
R &= \text{Diag}(0.5, 128, 128, 128).
\end{align}
For this problem, we use automatic differentiation to compute all required second derivatives for both IPOPT and CSVTO. We run IPOPT with two different maximum iteration settings. For the first, we limit the maximum number of iterations to 100 for the initial warm start and to 10 for subsequent time steps. We limit the number of iterations so that IPOPT has a comparable computation time to other baselines. The next setting is to set the maximum iterations to 1000, which allows IPOPT to run until convergence for most queries. We refer to this method as IPOPT-1000. As we will show in Section \ref{sec:comp_time}, this method is substantially slower than other baselines and prohibitively slow for MPC applications, but we included this baseline to see how well IPOPT performs when computation time is not an issue. For the baselines using a penalty method we use $\mu=2000$ and test two variants for $\lambda$: $\lambda=100$ and $\lambda=1000$. 

\begin{figure}
    \centering
    \includegraphics[width=0.49\textwidth]{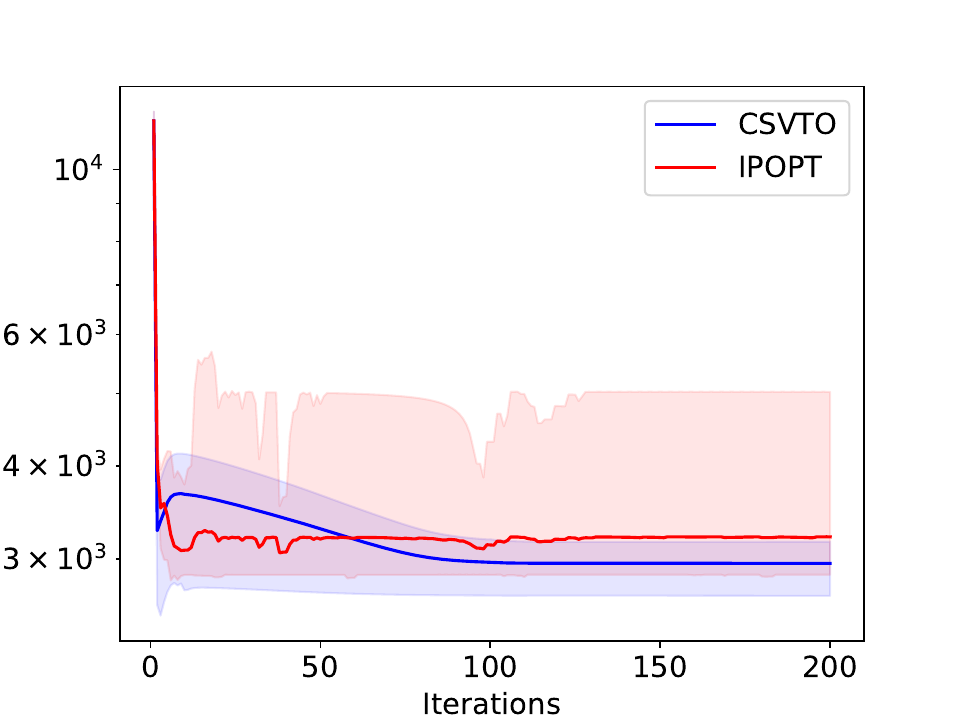}
    \caption{Comparison between CSVTO and IPOPT with multiple initializations on the quadrotor task with static obstacles. We compare CSVTO with 8 trajectory samples vs. 8 runs of IPOPT, both from the same initializations and record the minimum cost achieved from the 8 trajectories over 200 iterations of both. We run 10 trials for each method. The shaded regions show the range of the minimum cost achieved over the 10 trials. We see that from the same initializations, CSVTO finds a solution with a lower cost.}
    \label{fig:multi_starts}
\end{figure}

In Figure \ref{fig:multi_starts} we compare CSVTO and IPOPT run for 200 iterations for a single planning query with multiple different initializations, indicating that for the same initializations, CSVTO finds a lower cost local minimum. To generate these initializations, we sample a nominal control sequence from the prior $p(U)$ and use small Gaussian perturbations with $\sigma=0.01$ around this nominal control sequence as the initialization. The initial state sequence is found by applying these controls with the dynamics. We repeat this process 10 times for a different nominal control sequence. The results demonstrate that parallel trajectory optimization with CSVTO is beneficial even when the initial trajectory distribution is not diverse.

\begin{figure*}
    \centering
      \includegraphics[width=0.95\textwidth, trim=20 18 20 25,clip]{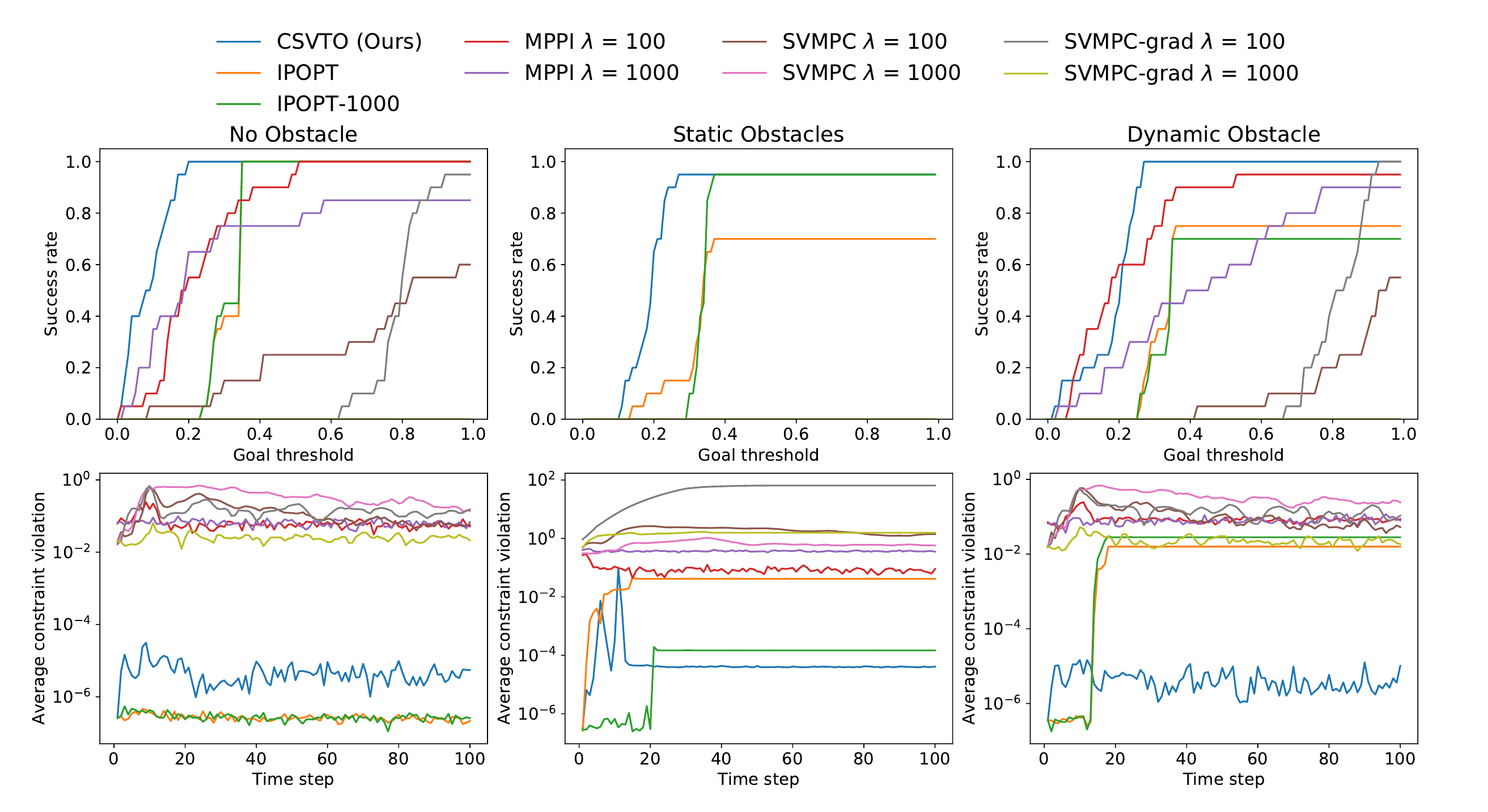}
    \caption{Results for quadrotor experiments. The top row shows the success rate as we increase the size of the goal region. The bottom row shows the average surface constraint violation as a function of time. Left) No obstacle. Middle) Static obstacles. Right) Dynamic obstacle.}
    \label{fig:results_quadrotor}
\end{figure*}

We ran the quadrotor experiments for the three different obstacle cases for 20 trials with randomly sampled starts. The results are shown in Figure \ref{fig:results_quadrotor}. CSVTO succeeds for 20/20 trials for the no obstacles and dynamic obstacles cases, and 19/20 for the static obstacle case, all with a goal threshold of 0.3m. For the static-obstacle and dynamic-obstacle experiments, IPOPT is the next best performing with 20/20 trials for no-obstacles at a goal threshold of 0.4m, but success falls to 15/20 for both the static-obstacles and dynamic-obstacle case. We see that running IPOPT with more iterations improves performance for the static obstacles case, but in the other two cases, there is no significant difference in performance when allowing IPOPT to run until convergence. However, running IPOPT to convergence has substantially higher computation time, which we will discuss further in section \ref{sec:comp_time}. For the no-obstacles and dynamic-obstacle cases, we see that sample-based methods perform well according to the task success rate, however, they fail to satisfy the surface equality constraint. In addition, both MPPI and SVMPC fail for the static obstacles case.

Trajectories generated from IPOPT vs CSVTO for the dynamic obstacles case are shown in Figure \ref{fig:quadrotor_exp}, IPOPT generates a trajectory aiming to go around the obstacle, but the movement of the obstacle renders that trajectory infeasible as time progresses. IPOPT is not able to adapt the trajectory to go around the obstacle. In contrast, CSVTO generates a multimodal set of trajectories that go either way around the obstacle. It is then able to update the trajectories effectively, avoiding the obstacle and reaching the goal. We do see that IPOPT achieves the lowest constraint violation in the case of no obstacle or a static obstacle, while CSVTO achieves the lowest constraint violation when there is a dynamic obstacle.

\subsection{Robot Manipulator on Surface}
\begin{figure}
    \centering
    \includegraphics[width=0.49\textwidth]{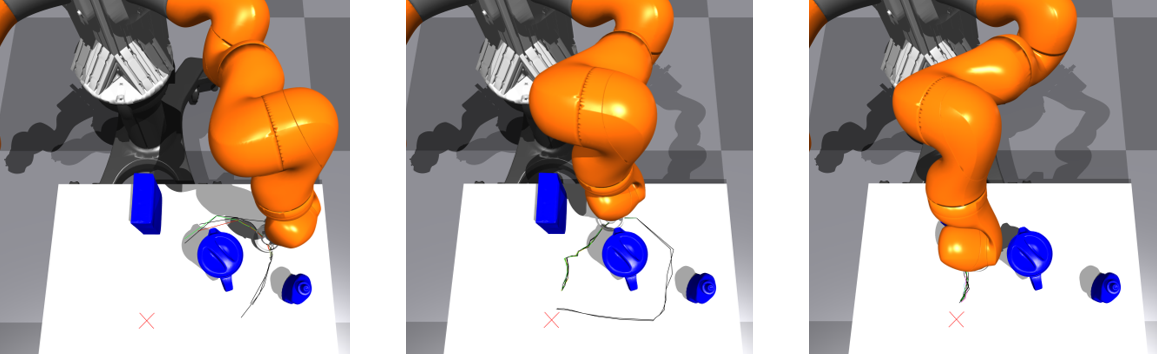}
    \caption{Snapshots from CSVTO used for the robot manipulator on a surface experiment. The robot must move the end-effector to a goal location while remaining on the surface of the table and avoiding the obstacles. CSVTO generates trajectories that explore different routes to the goal.}
    \label{fig:victor_table}
\end{figure}

In this task, we consider a 7DoF robot manipulator where the end effector is constrained to move in SE(2) along the surface of a table. The robot must move to a goal location while avoiding obstacles on the surface. The setup is shown in Figure \ref{fig:victor_table}. This system's state space is the robot's joint configuration $q \in \mathbb{R}^d$. The actions are the joint velocity $\dot{q}$ and the dynamics are given by Euler integration $q_{t+1} = q_t + \dot{q}_t \: dt$, with $dt=0.1$. The prior distribution over actions is $p(U)= \mathcal{N}(0, \sigma^2 I)$, where $\sigma = 0.5$. The planning horizon is 15. The cost is $C(\tau) = 2500 ||p_T^{xy} - p^{xy}_{goal}||_2 + 250 \sum^{T-1}_{t=1} ||p_t^{xy} - p^{xy}_{goal}||_2$, where $p^{xy}_t$ is the end effector $x,y$ position which is computed from the forward kinematics. The equality constraints on this system are  $p^{z}_t = 0.8$, which is the height of the table, and additionally, there is an orientation constraint that the z-axis of the robot end effector must be orthogonal to the table, i.e. the inner product of the table z-axis and the robot z-axis should be equal to -1.

While obeying the table constraint the robot must also avoid 3 obstacles from the Yale-CMU-Berkeley (YCB) object dataset \cite{calli2015ycb}. We enforce this with a constraint that the signed distance to the obstacles must be positive, which we compute from the meshes of the objects. Since signed distance functions (SDFs) are composable via the \texttt{min} operator, we combine the SDFs of the three obstacles into a single inequality constraint per timestep rather than an inequality constraint per obstacle. This is to reduce the total number of inequality constraints, as introducing more inequality constraints results in more slack variables and a higher dimensional problem. To evaluate this constraint, offline we generate points on the surface of the robot. Online, we use forward kinematics to map all of these points to the world frame and evaluate their SDF value, selecting the minimum SDF value as the value of the constraint. To compute the gradient of the constraint, consider that for any surface point we can compute the gradient of the SDF value with respect to the point from the object mesh. We then use automatic differentiation to backpropagate this gradient through the forward kinematics to compute a gradient of the SDF value with respect to the joint configuration. Finally, to calculate an overall gradient, we use a weighted combination of the gradients for each surface point, with the weight computed via a \texttt{softmin} operation on the SDF values.

The resulting inequality constraint is not twice differentiable, both because of non-smooth object geometries and because of composing SDFs with the \texttt{min} operator. Due to this, for CSVTO we omit the second-order term in equation (\ref{eq:grad_kernel}) for the inequality constraint, and for IPOPT we use L-BFGS to approximate second-order information. Computing the SDF value and gradient is a computationally expensive operation, so we pre-compute grids of the SDF values and the SDF gradients and do a look-up when performing the optimization. We use a $320 \times 320 \times 480$ grid with a resolution of $2.5mm$. There are also joint limit constraints on all of the robot joints. 

For the penalty-based baselines, we use penalty parameters of $\mu=2000$ and variants with $\lambda=100$ and $\lambda=1000$. For IPOPT, we found that running until convergence was prohibitively costly, taking several minutes to converge per optimization. For this reason, we limited the maximum number of iterations for IPOPT to be the same as CSVTO, resulting in a similar computation time. This is discussed further in Section \ref{sec:comp_time}.

Due to contact with the table, the dynamics of the system used for planning can deviate from those in the simulation. When computing the constraint violation, we use the actual constraint violation in the simulator rather than the planned constraint violation. 

\begin{figure}
    \centering
    \includegraphics[width=0.48\textwidth]{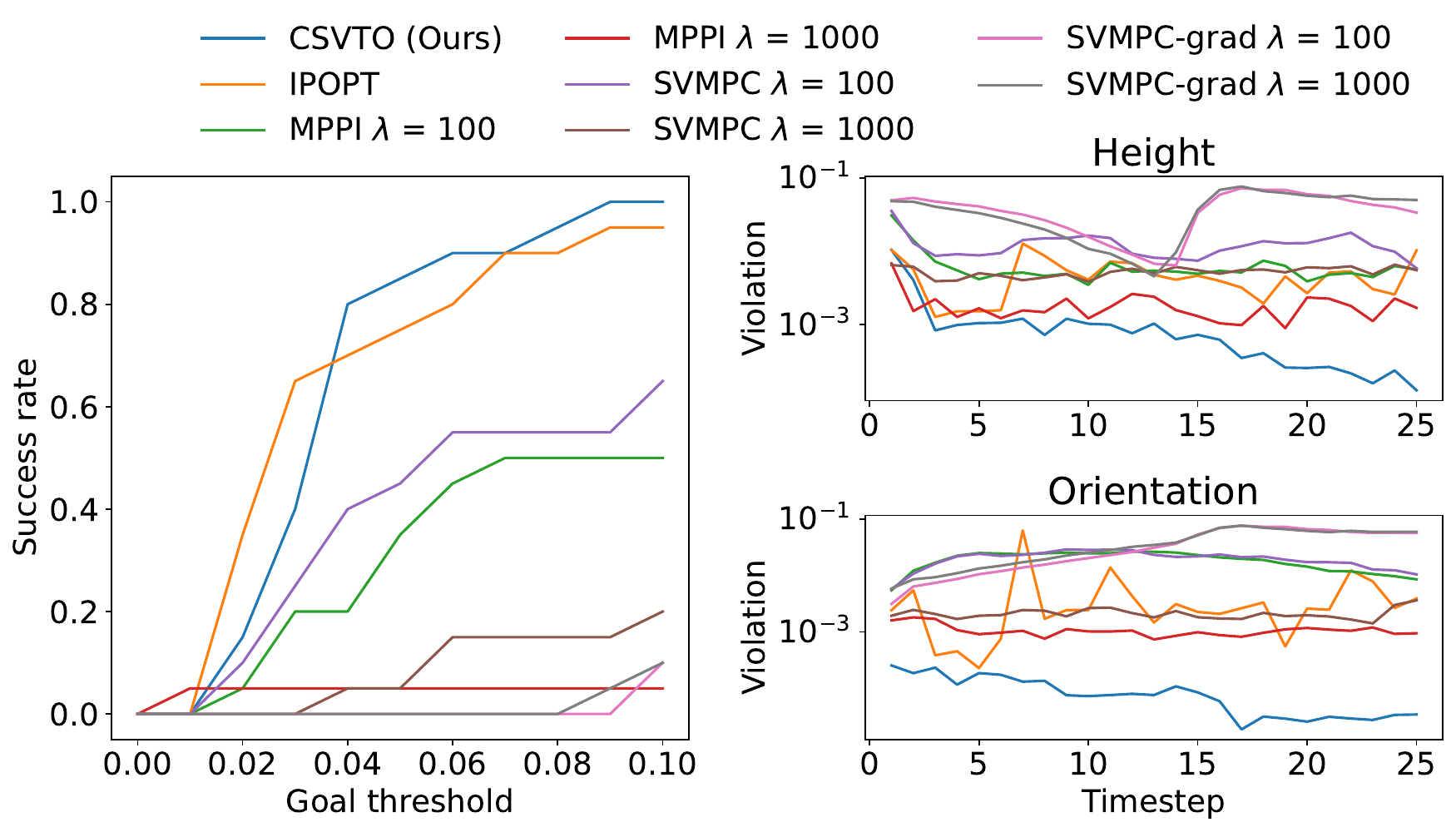}
    \caption{Results for robot manipulator on surface experiments Left column shows success rate as we increase the size of the goal region. Right column shows average constraint violation as a function of time for both the height constraint and the orientation constraint.}
    \label{fig:results_victor_table}
\end{figure}

We run this experiment for 20 trials with random goals and show the results in Figure \ref{fig:results_victor_table}. Our results show that CSVTO succeeds in all 20 trials with a goal threshold 0f 0.1m and achieves the lowest constraint violation of all methods. The next closest baseline, IPOPT succeeds 19/20 times, with the failure case resulting from a poor local minima with $q_t$ and $q_{t+1}$ on either side of an obstacle, but a large distance from one another. This resulted in the robot becoming stuck on the obstacle and unable to make progress.

\begin{table*}
\caption{Hyperparameter values for the three experiments}
\centering
\label{table:hyperparameters}
\begin{tabular}{ c|c|c|c|c|c|c|c|c|c|c|c } 
Experiment & \# particles & $\alpha_J$ & $\alpha_C$ & $\epsilon$ & $K_{w}$ & $K_{o}$ & \texttt{resample\_steps} & $\beta$ & $\sigma_{resample}$ & $\lambda$ & $W$\\
\hline
Quadrotor & 8 & 0.05 & 1 & 0.5 & 100 & 10 & 10 & 0.55 & 0.1 & 1000 & 3\\
Manipulator on Surface &  8 & 0.01 & 1 & 0.1 & 100 & 10 & 1 & 0.1 & 0.01 & 1000 & 3\\
Manipulator wrench & 4 & 0.01 & 1 & 0.25 & 100 & 10 & 1 & 0.1 & 0.01 & 1000 & 3\\
\end{tabular}
\end{table*}

\subsection{Robot Manipulator using wrench}
In this task, we consider a 7DoF robot manipulator in which the goal is to manipulate a wrench to a goal angle. To turn the wrench, the robot must be able to supply at least 1Nm of torque. The setup is shown in Figure \ref{fig:victor_wrench}. The state space is $[q \quad \phi \quad \theta]^T$. $q \in \mathbb{R}^7$ is the configuration space of the robot. $\phi$ parameterizes the distance between the robot end-effector and the wrench in the x-y plane as $l + \phi$ where $l$ is a nominal distance. $\theta$ is the wrench angle. The actions are the joint velocities $\dot{q}$. The dynamics of the joint configuration are given by Euler integration $q_{t+1} = q_t + \dot{q}_t \: dt$, with $dt=0.1$. We use a simple geometric model for dynamics of $\phi$ and $\theta$. Assuming that the robot end-effector remains grasping the wrench we compute the next $\phi$ as $\phi = ||p^{xy}_{ee} - p^{xy}_{wrench}||_2 - l$. To compute the next joint angle $\theta$ we use $\theta_{t+1} = \theta_{t} + \tan \frac{\Delta x_{ee}}{\Delta y_{ee}}$. The prior distribution over actions is $p(U)= \mathcal{N}(0, \sigma^2 I)$, where $\sigma = 1$. 

The equality constraints of the system are that $p^z_{ee}$ should be at a fixed height, and additionally that $\theta_T = \theta_{goal}$.  There is also a constraint that the end-effector orientation of the robot remains fixed relative to the wrench. To do this we compute the desired end-effector orientation from the wrench angle, and compute the relative rotation between the desired and actual end-effector orientation in axis-angle form, constraining the angle to be zero. In total, combining the dynamics constraints for $\phi$ and $\theta$ with the other equality constraints there are four equality constraints on the pose of the end-effector per time step. When reporting the constraint violation, we report the maximum violation of these four constraints. The inequality constraints of the system are that the desired torque should be achievable within the robot joint limits. This constraint is $\texttt{min\_torque} \leq J(q)^T(l + \phi) \leq \texttt{max\_torque}$ where $J$ is the manipulator jacobian. There are also joint limit bound constraints, and a bound constraint on $\phi$. Computing the second derivative of this constraint requires computing the second derivative of the manipulator Jacobian, which is costly. To avoid this, for CSVTO we omit the second-order terms in equation (\ref{eq:grad_kernel}) and for IPOPT we use L-BFGS. There is no cost $C$ for this experiment, instead, the inference problem reduces to conditioning the prior on constraint satisfaction. The planning horizon is 12.

For the penalty-based methods we use $\mu=1000$ and variants with $\lambda=1000$ and $\lambda=10000$. We run IPOPT both until convergence with a max number of iterations of 1000 and additionally with a max iterations of 200 at warmup and 20 online, which results in a similar computation time to CSVTO.

\begin{figure}
    \centering
    \includegraphics[width=0.48\textwidth]{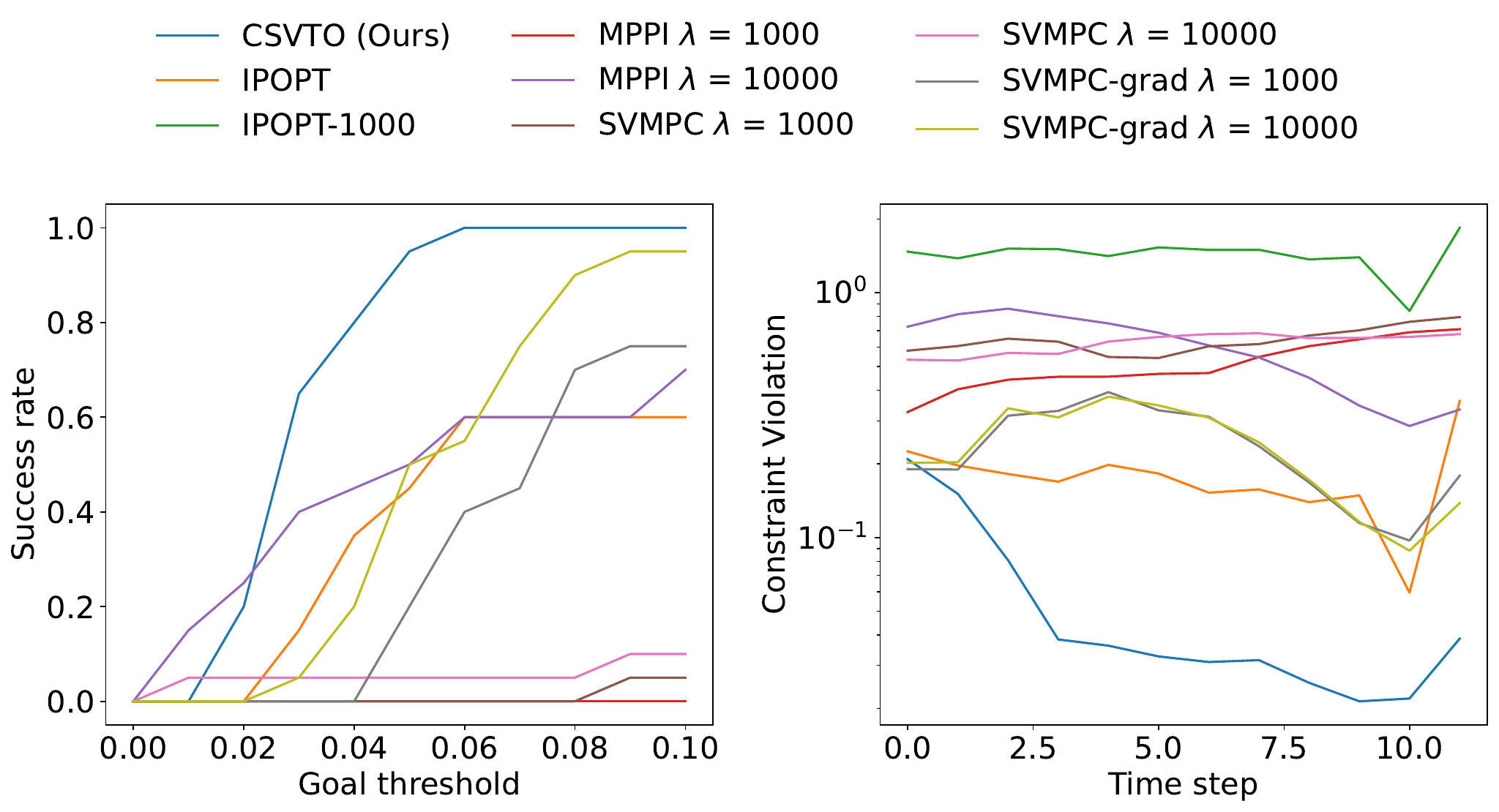}
    \caption{Results for the robot manipulator using the wrench. Left column shows the success rate as we increase the size of the goal region. The right column shows the average constraint violation as a function of time time, where we compute the constraint violation at a given time via the maximum violation among the equality constraints. }
    \label{fig:results_victor_wrench}
\end{figure}

We run this experiment for 20 trials with random initializations and show the results in the bottom row of Figure \ref{fig:results_victor_wrench}. This problem is challenging because the dynamics are based on a simple inaccurate geometric model. Compliance in the gripper causes deviation from this geometric model, and the model is only accurate so long as all constraints hold. Our results show that CSVTO can succeed in all 20 trials with a goal threshold of 0.06 radians and achieves the lowest constraint violation. The next closest baseline, SVMPC-grad with $\lambda = 10000$ succeeds 19/20 times with a goal threshold of 0.09 radians, dropping to 11/20 at 0.06 radians. We find that running IPOPT to convergence leads to poor performance, as the solver is unable to converge to a feasible solution. Limiting the maximum iterations to 200 for the initial warm-start and 20 for subsequent online iterations leads to improved task performance, achieving a success rate of 12/20.
  
We also demonstrate CSVTO on real hardware for the robot manipulator manipulating a wrench task, shown in Figure \ref{fig:victor_real}. After generating a configuration-space trajectory using CSVTO, we command the robot to move to the first configuration waypoint of that trajectory using a joint impedance controller. Once the robot has reached the desired waypoint, we perform re-planning to generate a new configuration-space trajectory. We use the same hyperparameters as those in the simulator for this experiment. During execution, we applied disturbances by perturbing the robot end-effector. The impedance controller can reject small disturbances, but larger disturbances require re-planning from the perturbed location. Figure \ref{fig:victor_real} shows one such perturbation. Despite large disturbances, our method was able to readjust the grasp and complete the task successfully.

\subsection{Computation Time}
\label{sec:comp_time}
To determine the computation times for CSVTO and each baseline, we ran 10 trials for each experiment on a computer with an Intel i9-11900KF Processor with an NVIDIA RTX 3090 GPU. We record the average computation times for the initial trajectory as well as subsequent online trajectories, which we refer to as $t_w$ and $t_o$, respectively. We also record the standard deviations of the computation times. The number of iterations used for the warm-up and online phase is $K_w$ and $K_o$, respectively. For IPOPT this is a maximum number of iterations, and the solver may terminate early. For all other methods, all iterations are used. 

\subsubsection{12DoF Quadrotor}
The average computation time of CSVTO compared to baselines for all quadrotor experiments is shown in Table 
\ref{table:quadrotor_computation_times}. For this experiment, computing the gradient was a major computational bottleneck, thus for the sample-based methods we allowed them more iterations. We see that MPPI and SVMPC are faster than CSVTO with online trajectory computation times of $0.366s$, $0.439s$, and $0.589s$, respectively. For the no-obstacles and dynamic-obstacle cases, IPOPT is also faster than CSVTO with an average online computation time of $0.429s$ and $0.479s$ due to early termination. However, for the static obstacles case, this rises to $0.768s$ compared to CSVTO at $0.650s$. When running IPOPT to convergence, the solving time is substantially larger, with an average computation time for the static obstacle case of $15.8s$. We also see that the standard deviations are very large, due to the variability in how quickly the solver converges. Combining these with the results from Section \ref{sec:eval_quadrotor}, we see that CSVTO outperforms IPOPT to convergence with substantially faster computation times. 

\subsubsection{Robot 7DoF Manipulator}
The computation times for all methods on both 7DoF manipulation experiments are shown in Table \ref{table:victor_computation_times}. For the manipulator on a surface experiment, the difference in computation speed of the sample-based vs gradient-based algorithms per iteration was less pronounced than for the quadrotor experiment. We thus kept the number of iterations the same for all experiments, with 100 warm-up iterations and 10 online iterations. CSVTO and IPOPT have similar computation times at $1.12s$ and $1.14s$ to compute a trajectory online. MPPI is again the fastest algorithm at $0.691s$ to generate a trajectory online, though the performance is lower both in terms of task success and constraint violation. Initial attempts to run IPOPT with a maximum of 1000 iterations took several minutes to solve, which rendered it impractical.

For the wrench task, CSVTO and SVMPC-grad have similar computation times. While CSVTO requires the computation of the second derivative of the constraints, the cost evaluation of SVMPC-grad requires a loop through the time horizon, slowing down both cost and gradient evaluation. Since CSVTO employs a collocation scheme this process is vectorized. Whether CSVTO or SVMPC-grad is faster depends on the relative cost of computing the second derivatives vs. looping through the time horizon. Each iteration of IPOPT was faster than CSVTO for this experiment, as IPOPT using the L-BFGS approximation computes no second derivatives, whereas CSVTO only neglected the second derivatives of the force inequality constraint. We thus allowed IPOPT more iterations, as seen in Table \ref{table:victor_computation_times}. Attempting to allow IPOPT to run with a much larger maximum iteration number resulted in much slower solving times and worse performance. 


\begin{table*}
\caption{Mean and standard deviation of computation times for CSVTO and all baseline methods for the 12DoF quadrotor experiments. $t_w$ and $t_o$ are the average times taken to generate the trajectories for the warm-up phase and online phase, respectively}
\begin{center}
\begin{tabular}{ c|c|c|c|c|c|c|c|c } 
& & & \multicolumn{2}{c|}{No Obstacles} & \multicolumn{2}{c|}{Static Obstacles} & \multicolumn{2}{c}{Dynamic Obstacle}  \\
\hline
Method & $K_{w}$ & $K_{o}$ & $t_w$ (s) & $t_o$ (s) & $t_w$ (s) & $t_o$ (s)& $t_w$ (s) & $t_o$  (s) \\
\hline
CSVTO (Ours) & 100 & 10 & $5.92 \pm 0.235$ & $0.589 \pm 0.003$ & $6.56 \pm 0.39$ & $0.650 \pm 0.025$ & $6.47 \pm 0.344$ &  $0.643 \pm 0.021$   \\
IPOPT & 100 & 10 & $4.36 \pm 2.29$  & $0.429 \pm 0.008$ & $7.19 \pm 2.48$ & $0.768 \pm 0.069$ & $3.19 \pm 1.89$ &  $0.479 \pm 0.097$ \\
IPOPT-1000 & 1000 & 1000 & $17.5 \pm 30.2$ & $2.40 \pm 1.10$ & $39.2 \pm 32.0$ & $15.8 \pm 10.1$ & $10.8 \pm 24.5$ & $2.45 \pm 2.26$ \\
SVMPC-grad & 100 & 10 & $8.25 \pm 0.080$ & $0.771 \pm 0.217$ & $8.24 \pm 0.061$& $0.765 \pm 0.23$ & $8.28 \pm 0.054$ & $0.850 \pm 0.014$ \\
SVMPC & 250 & 25 & $4.26 \pm 0.030$ & $0.439 \pm 0.002$ & $6.07 \pm 0.031$ & $0.621 \pm 0.003$ & $4.35 \pm 0.24$ & $0.449 \pm 0.017$ \\
MPPI & 250 & 25 & $3.63 \pm 0.021$ & $0.366 \pm 0.0019$ & $5.45 \pm 0.039$ & $0.55 \pm 0.003$ & $3.66 \pm 0.13$& $0.373 \pm 0.016$  \\
\end{tabular}
\label{table:quadrotor_computation_times}
\end{center}
\end{table*}

\begin{table*}
\caption{Average computation times for CSVTO and all baseline methods for the 7DoF robot manipulator experiments. $t_w$ and $t_o$ are the average times taken to generate the trajectories for the warm-up phase and online phase, respectively}
\begin{center}
\begin{tabular}{ c|c|c|c|c|c|c|c|c } 
& \multicolumn{4}{c|}{Surface} & \multicolumn{4}{c}{Wrench}  \\
\hline
Method & $K_{w}$ & $K_{o}$ & $t_w$ (s) & $t_o$ (s)& $K_{w}$ & $K_{o}$ & $t_w$ (s) & $t_o$ \\
\hline
CSVTO (Ours) & 100 & 10 & $9.41 \pm 0.42$ & $1.12 \pm 0.19$ & 100 & 10 & $9.62 \pm 0.84$ & $0.64 \pm 0.004$ \\
IPOPT & 100 & 10 & $10.26 \pm 3.5$ & $1.14 \pm 0.27$ & 200 & 20 & $5.82 \pm 0.54$ & $0.493 \pm 0.028$ \\
IPOPT-1000 & 1000 & 1000 & --- & --- & 1000 & 1000 & $30.8 \pm 2.51$ & $22.7 \pm 2.84$ \\
SVMPC-grad & 100 & 10 & $8.55 \pm 0.072$ & $1.10 \pm 0.27$ & 100 & 10 & $9.54 \pm 0.071$ & $0.732 \pm 0.004$ \\
SVMPC & 100 & 10 & $7.27 \pm 0.097$ & $0.758 \pm 0.010$ & 100 & 10 & $7.44 \pm 0.15$ & $0.571 \pm 0.007$ \\
MPPI & 100 & 10 & $6.91 \pm 0.12$ & $0.691 \pm 0.028$ & 100 & 10 & $7.05 \pm 0.11$ & $0.506 \pm 0.006$ \\
\end{tabular}
\label{table:victor_computation_times}
\end{center}
\end{table*}

\begin{figure}
    \centering
    \includegraphics[width=0.49\textwidth]{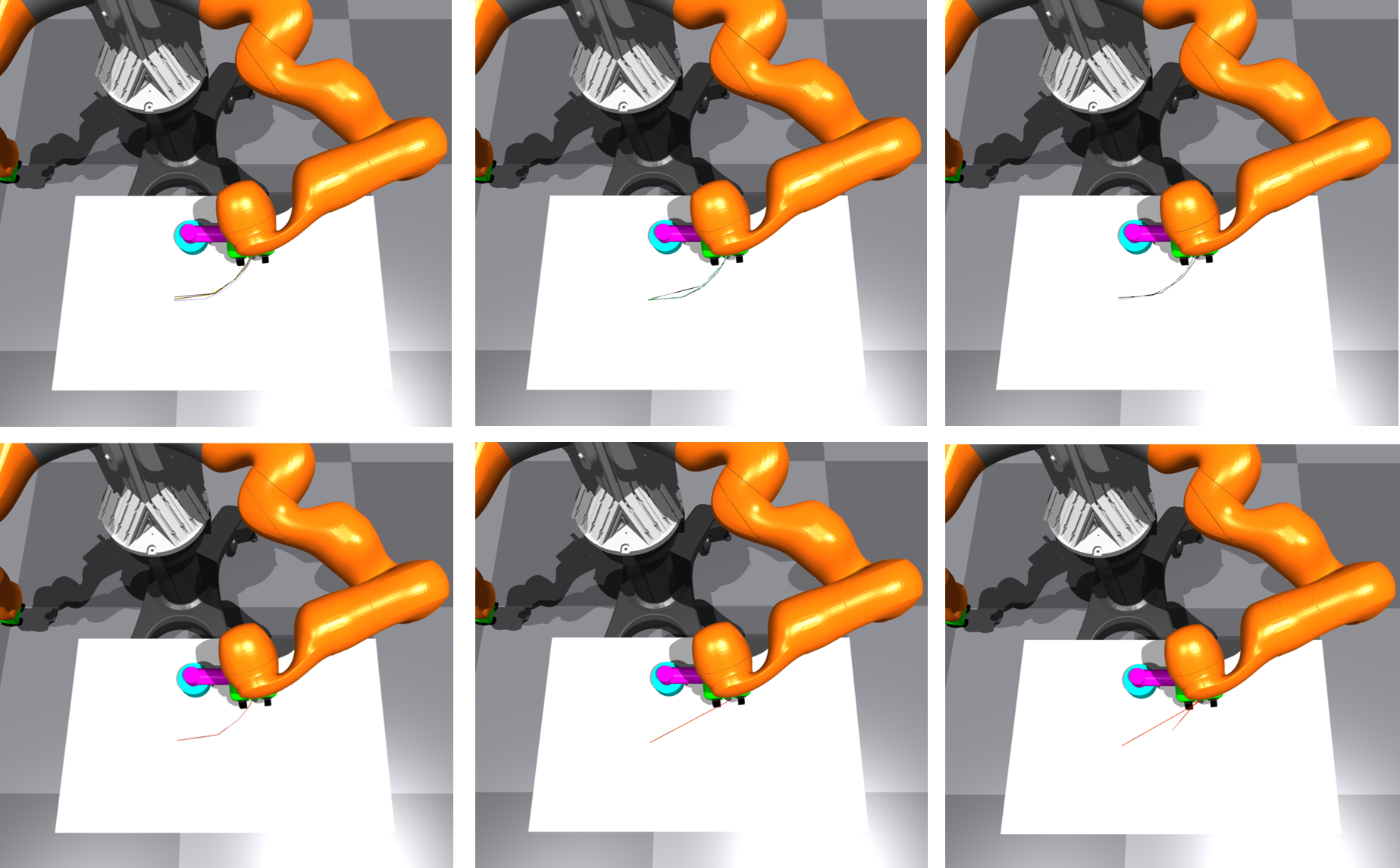}
    \caption{The robot manipulator turning a wrench experimental set-up. The goal is to turn the wrench by 90 degrees. End-effector planned path at the first time-step visualized for three different initial trajectories generated by  CSVTO (Top) and IPOPT (Bottom). CSVTO's end-effector path traces an arc around the wrench center to turn the wrench, while IPOPT paths are often poor, containing very large steps and lacking smoothness}.
    \label{fig:victor_wrench}
\end{figure}

\section{Discussion}
In this section we will discuss some of the advantages of CSVTO over baselines, and then discuss some limitations and finally highlight areas for future work.
\subsection{Local minima}
CSVTO produces diverse approximately constraint-satisfying trajectories. By encouraging diversity through the course of the optimization the algorithm searches the solution space more widely and can result in multi-modal sets of solutions, for example, see Figure \ref{fig:quadrotor_exp}. We found that this behavior is beneficial for escaping from local minima. This was most clearly demonstrated in the 12DoF Quadrotor experiment. We found that in the case of no obstacles, IPOPT was consistently able to get relatively close to the goal, achieving a 100\% success rate at a goal region of 0.4m. However, it was unable to escape a local minimum in the vicinity of the goal region. This local minimum appears to be induced by the surface constraint, as IPOPT frequently became stuck at a position where it needed to climb in height to reach the goal while satisfying the constraint, incurring a large control cost. In contrast, CSVTO was able to achieve a 100\% success with a much smaller goal region of 0.2m.
\subsection{Initialization}
CSVTO optimizes a set of trajectories in parallel. Each of these trajectories has a different random initialization, and, as mentioned, the objective encourages trajectory diversity. We find that this approach is effective at making the algorithm more robust to poor initialization. This is most clearly seen in the 7DoF wrench manipulation experiment, shown in Figure \ref{fig:victor_wrench}. This system is highly constrained, and we can see from figure \ref{fig:victor_wrench} that the trajectories generated by IPOPT can be very low quality when poorly initialized. This is reflected in the success rates, where in our experiments CSVTO succeeds for 20/20 of the trials vs. 12/20 for IPOPT. 
\subsection{Limitations \& Future Work}
\paragraph{Differentiability} Our method requires that all costs and constraints are differentiable. This is a restrictive assumption, particularly when treating dynamics as a constraint. Many contact-rich robot manipulation tasks exhibit discontinuities that invalidate this assumption.
\paragraph{Slack variables} Our approach converts inequality constraints to equality constraints by introducing slack variables. While this is a natural way of incorporating inequality constraints into our method, it results in increasing the number of decision variables by the number of inequality constraints. This is likely to be problematic for long-horizon planning tasks with many inequality constraints. A possible solution would be solving a QP subproblem at every iteration to determine the active inequality constraints as in \cite{feppon2020optim}, however, this has the issue that we would need to solve an individual QP subproblem for every particle. 
\paragraph{Computation time inadequate for real-time control} We note from Table \ref{table:quadrotor_computation_times}, in the dynamic obstacle quadrotor task the average computation time for online trajectory generation is $0.643s$ for CSVTO, compared to MPPI, the fastest baseline, taking $0.373s$. In this case, the solve times for the current implementation of CSVTO and all baselines are insufficient for real-time reactive control. Our method, all baselines other than IPOPT, and all cost and constraint functions were implemented in Python, using automatic differentiation in PyTorch to compute the relevant first and second derivatives. Implementing these methods in C++, using a library such as CasADI \cite{Andersson2019} for automatic differentiation, may enable real-time performance on these systems in future work.

\paragraph{Kernel selection} While our approach decomposes the kernel into a sum of kernels operating on sub-trajectories, each of these kernels is an RBF kernel. While the RBF has attractive properties, such as strict positive-definiteness and smoothness, we believe that exploring task-specific kernels for trajectory optimization is an interesting avenue for future work. 

\section{Conclusion}
In this article, we presented Constrained Stein Variational Trajectory Optimization (CSVTO), an algorithm for performing constrained trajectory optimization on a set of trajectories in parallel. To develop CSVTO we formulated constrained trajectory optimization as a Bayesian inference problem, and proposed a constrained Stein Variational Gradient Descent (SVGD) algorithm inspired by O-SVGD \cite{constrained_stein} for approximating the posterior over trajectories with a set of particles. Our results demonstrate that CSVTO outperforms baselines in challenging highly-constrained tasks, such as a 7DoF wrench manipulation task, where CSVTO succeeds in 20/20 trials vs 12/20 for IPOPT. Additionally, our results demonstrate that generating diverse constraint-satisfying trajectories improves robustness to disturbances, such as changes in the environment, as well as robustness to initialization.

\appendix[Matrix Derivative of \texorpdfstring{$P(\tau)$}{Projection}]
\label{sec:appendix}
In Equation (\ref{eq:grad_kernel}) we showed that the repulsive gradient is split into two terms, one of which contains the matrix derivative $\nabla_{[\tau]_k} P(\tau)$. In this section, we show how to compute this derivative. For notational convenience, let $\tau \in \mathbb{R}^N$ (thus $P(\tau) \in \mathbb{R}^{N \times N}$), $h(\tau) \in \mathbb{R}^M$ (where $M$ is the number of constraints), and we omit the dependence on $\tau$ when writing the constraint derivative $\nabla h(\tau)$.
$\nabla_{[\tau]_k} P(\tau)$ is a matrix of shape $N \times N$. 
We refer to the second derivative of the $l$th constraint $\nabla^2 h_l(\tau)$ as $H_l$, which is an $N \times N$ matrix. The matrix derivative $\nabla_{[\tau]_k} P(\tau)$, as defined in Equation (\ref{eq:grad_kernel}), can be expanded into three terms:
\begin{equation}
\nabla_{[\tau]_k} [P(\tau)]_{i,k} = 2 A_{i, k} - B_{i, k},
\end{equation}
where $A, B \in \mathbb{R}^{N \times N}$ and $i,k \in \{1, ..., N\}$. $A_{i, k}$ is given by
\begin{equation}
    A_{i, k} =  \sum^M_l [H_l]_{k, i}[\left (\nabla h \nabla h^T\right)^{-1} \nabla h]_{l, k}.
\end{equation}
To compute $B_{i,k}$, we first consider the matrix $D_k \in \mathbb{R}^{M \times M}$:
\begin{equation}
    [D_k]_{l, m} = \sum^N_j \left( [H_l]_{k, j} [\nabla h]_{l, j} + [H_m]_{j, k} [\nabla h]_{m, j} \right),
\end{equation}
\noindent for $l,m \in \{1, ...,M\}$. We then finally compute $B_{i,k}$ as
\begin{equation}    
\begin{split}
    & B_{i, k} = \\ 
    & \sum^M_{l} \sum^M_m [D_k]_{l, m} [\nabla h^T\left (\nabla h \nabla h^T \right)^{-1}]_{i, l} [\left (\nabla h \nabla h^T\right)^{-1} \nabla h]_{m, k}.
\end{split}
\end{equation}
When neglecting second-order terms for the $l$th constraint $h_l(\tau)$ (as discussed in Section \ref{sec:method:repulsion}), we set $H_l = \mathbf{0}$ when computing $A_{i, k}$ and $B_{i, k}$. 
\bibliographystyle{IEEEtran}
\bibliography{references}
\vspace{2cm}
\begin{IEEEbiography}
[{\includegraphics[width=1in,height=1.25in,clip,keepaspectratio]{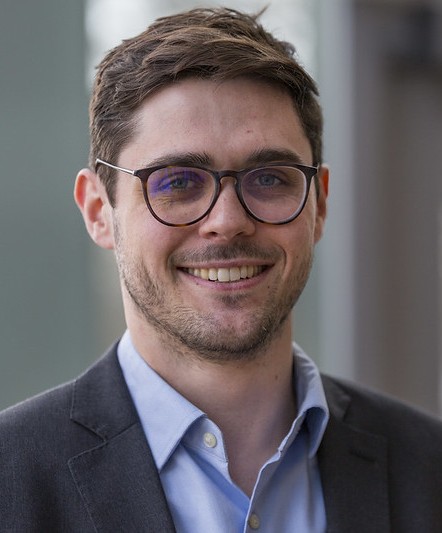}}]{Thomas Power}
received the M.Eng. degree in mechanical engineering from Imperial College London, London, UK, in 2016, and the M.S. and Ph.D. degrees in Robotics from the University of Michigan, Ann Arbor, MI, USA, in 2020 and 2024 respectively. His current research interests include trajectory optimization and machine learning applied to robotic manipulation.
\end{IEEEbiography}
\begin{IEEEbiography}
[{\includegraphics[width=1in,height=1.25in,clip,keepaspectratio]{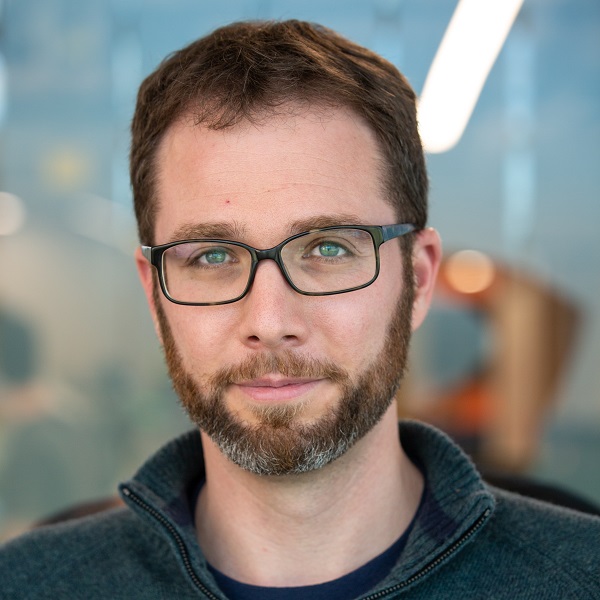}}]{Dmitry Berenson} is an Associate Professor in the Robotics Department at the University of Michigan, where he has been since 2016. Before coming to University of Michigan, he was an Assistant Professor at WPI (2012-2016). He received a BS in Electrical Engineering from Cornell University in 2005 and received his Ph.D. degree in robotics from the Robotics Institute at Carnegie Mellon University in 2011. He was also a post-doc at UC Berkeley (2011-2012). He has received the IEEE RAS Early Career Award and the NSF CAREER award. His current research focuses on robotic manipulation, robot learning, and motion planning.
\end{IEEEbiography}
\end{document}

%% file: terms.tex
\newtheorem{thm}{Theorem}